\documentclass{article}
\PassOptionsToPackage{numbers, compress}{natbib}
   \usepackage[final]{neurips_2023}

\usepackage{graphicx}
\usepackage[outdir=./]{epstopdf}
\usepackage[utf8]{inputenc} % allow utf-8 input
\usepackage[T1]{fontenc}    % use 8-bit T1 fonts
\usepackage{hyperref}       % hyperlinks
\usepackage{url}            % simple URL typesetting
\usepackage{booktabs}       % professional-quality tables
\usepackage{amsfonts}       % blackboard math symbols
\usepackage{nicefrac}       % compact symbols for 1/2, etc.
\usepackage{microtype}      % microtypography
\usepackage{xcolor}         % colors
\usepackage{color}
\usepackage{array}
\usepackage[linesnumbered,ruled,noend]{algorithm2e}
\usepackage{enumitem}
\usepackage{wrapfig}
\usepackage{amsmath}
\usepackage{tabu}                     % 表格插入
\usepackage{multirow}                 % 一般用以设计表格，将所行合并
\usepackage{multicol}                 % 合并多列
\usepackage{multirow}                % 合并多行
\usepackage{float}                    % 图片浮动
\usepackage{makecell}                 % 三线表-竖线
\usepackage{caption}
\usepackage{subcaption}
\usepackage{comment}

\newcommand{\eg}{\textit{e}.\textit{g}.}

\title{IPMix: Label-Preserving Data Augmentation Method
for Training Robust Classifiers}

\usepackage{authblk}
\usepackage{xpatch}
\xpatchcmd{\author}{\relax#1\relax}{\relax\detokenize{#1}\relax}{}{}
\author[\empty]{\textbf{Zhenglin Huang}\textsuperscript{$1$}}
\author[\empty]{\textbf{Xianan Bao}\textsuperscript{$1$}}
\author[\empty]{\textbf{Na Zhang}\textsuperscript{$1$}\thanks{Corresponding author}}
\author[\empty]{\textbf{Qingqi Zhang}\textsuperscript{$2$}}
\author[\empty]{\\\textbf{Xiaomei Tu}\textsuperscript{$3$}}
\author[\empty]{\textbf{Biao Wu}\textsuperscript{$1$}}
\author[\empty]{\textbf{Xi Yang}\textsuperscript{$4$}}
\affil[$1$]{School of Artificial Intellienge, Zhejiang Sci-Tech University}
\affil[$2$]{Yamaguchi University \hspace{0.3pt} \textsuperscript{$3$}ZGUC \hspace{0.3pt}} 
\affil[$4$]{University of Science and Technology of China, Hefei, China}
\begin{document}

\maketitle

\begin{abstract}
Data augmentation has been proven effective for training high-accuracy convolutional neural network classifiers by preventing overfitting. However, building deep neural networks in real-world scenarios requires not only high accuracy on clean data but also robustness when data distributions shift. While prior methods have proposed that there is a trade-off between accuracy and robustness, we propose IPMix, a simple data augmentation approach to improve robustness without hurting clean accuracy. IPMix integrates three levels of data augmentation (image-level, patch-level, and pixel-level) into a coherent and label-preserving technique to increase the diversity of training data with limited computational overhead. To further improve the robustness, IPMix introduces structural complexity at different levels to generate more diverse images and adopts the random mixing method for multi-scale information fusion. Experiments demonstrate that IPMix outperforms state-of-the-art corruption robustness on CIFAR-C and ImageNet-C. In addition, we show that IPMix also significantly improves the other safety measures, including robustness to adversarial perturbations, calibration, prediction consistency, and anomaly detection, achieving state-of-the-art or comparable results on several benchmarks. Code is available at {\url{https://github.com/hzlsaber/IPMix}}. 

\end{abstract}

\section{Introduction}
Deep neural network models have recently achieved remarkable performance on various computer vision tasks, such as zero-shot image classification~\cite{Wang2018ZeroShotIC,Zhai2021LiTZT,Gu2021OpenvocabularyOD}, 3D object detection~\cite{Maturana2015VoxNetA3,Yang2018PIXORR3,Mao2021VoxelTF}, and face recognition~\cite{Wang2017NormFaceLH,Wang2018CosFaceLM}. In real-world scenarios, models can achieve impressive accuracy when training and test distributions are identical, but challenges appear when confronted with out-of-distribution examples~\cite{Recht2019DoIC,Croce2020RobustBenchAS,Hendrycks2021UnsolvedPI}, such as natural corruptions~\cite{Mintun2021OnIB}, adversarial perturbations~\cite{Metzen2017OnDA}, and anomaly patterns~\cite{Elsayed2020NetworkAD}, necessitating robustness across distribution shifts. Data augmentation (DA) has been proposed to partially alleviate this issue, which applies diverse transformations on clean images to generate new training examples~\cite{Gong2020KeepAugmentAS,Calian2021DefendingAI}. Furthermore, a high diversity of augmented images enables neural networks to resist data distribution shifts and improve robustness~\cite{Subbaswamy2021EvaluatingMR}. DA approaches generally fall into three subgroups: image-level, patch-level, and pixel-level augmentations. 

Image-level augmentation techniques~\cite{Lim2019FastA,Cubuk2019AutoAugmentLA,Cubuk2019RandaugmentPA} apply transformations on the whole image, such as brightness, sharpness, and solarization, to increase the total amount of training data. Patch-level augmentation techniques~\cite{Takahashi2018RICAPRI,Kim2021CoMixupSG} typically mask or replace a region of an image, compelling classifiers to focus on less discriminative portions. Meanwhile, pixel-level augmentation techniques~\cite{Zhang2017mixupBE, Hendrycks2021PixMixDP} mix images using pixel-wise weighted averages to increase diversity within the training dataset.
%\textbf{Image-level augmentation techniques} apply transformations on the whole image, such as brightness, sharpness, and solarization, to increase the total amount of training data. Some image-level data augmentation techniques employ diverse combinations of transformations to increase the diversity of the training data, which can increase the chance the model becomes invariant to natural corruptions, improving the robustness of out-of-distribution.

%\textbf{Pixel-level augmentation techniques} mix images using pixel-wise weighted averages to increase diversity within the training dataset, leading to improved model generalization. One founding example is MixUp, which generates augmented images by linearly interpolating between two randomly selected images and their corresponding labels, providing a smoother decision boundary for the model. 

%\textbf{Patch-level augmentation techniques} typically mask or replace a region of an image, compelling classifiers to focus on less discriminative portions. Some of these methods utilize masks, additive Gaussian noise, or saliency maps to create partially occluded data outside the training distribution, thereby enhancing robustness. 

Previous studies have focused on either pixel-level or patch-level information to improve model performance. However, most of these techniques are label-variant, which may lead to manifold intrusion~\cite{Guo2018MixUpAL,Baena2022PreventingMI} and decrease performance on unseen data. Simultaneously, a limitation of image-level data augmentation techniques is the computationally expensive search for an optimal augmentation policy, often exceeding the training process's complexity~\cite{Lim2019FastA,Cubuk2019AutoAugmentLA}. Given these considerations and the potential for enhancing data augmentation strategies, we mainly discuss one question in this paper:  \textbf{\emph{How to take advantage of the strengths of the three methods while avoiding their drawbacks?}} 

\textbf{Our contributions are as follows:}

\begin{itemize}[leftmargin=0.5cm, itemindent=0cm]
    \item  We propose \textbf{IPMix},  a label-preserving data augmentation approach, which integrates three levels of data augmentation into a single framework with limited computational overhead, demonstrating that these approaches are complementary and that a unification among them is necessary to achieve robustness.
    \item  To further enhance model performance, IPMix incorporates structural complexity from synthetic data at various levels to produce more diverse images. Additionally, we employ random mixing methods and scar-like image patches for multi-scale information fusion.
    \item Extensive experiments demonstrate that IPMix achieves state-of-the-art corruption robustness and improves numerous safety metrics compared with other data augmentation approaches.
\end{itemize}

\begin{figure}[tb]
\vspace{-2.5em}
\begin{center}
\centerline{\includegraphics[width=1\columnwidth]{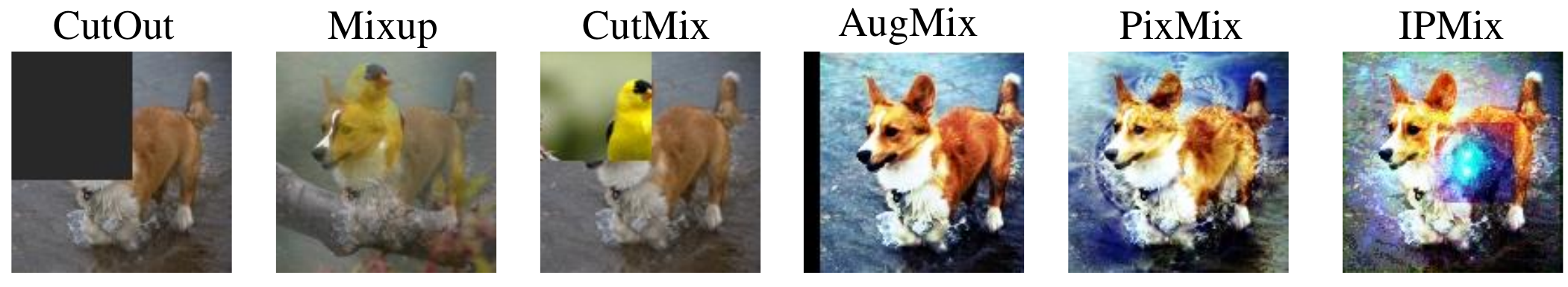}}
\caption{Visual comparison of various data augmentation methods. IPMix utilizes the structural complexity of fractals and multi-scale information to generate more diverse examples.}
\label{visual IPMix}
\end{center}
\vspace{-2.2em}
\end{figure}

\begin{wrapfigure}{r}{7cm}
\centering
\includegraphics[width=0.43\textwidth]{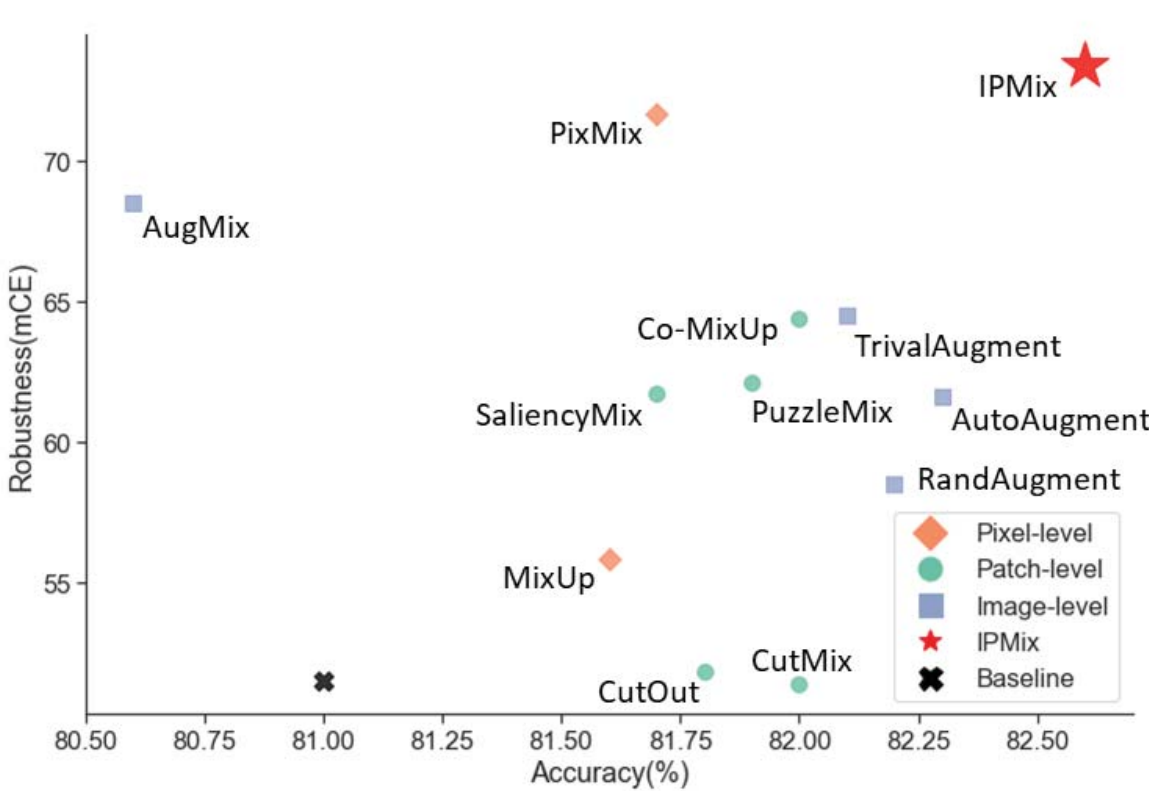}
\caption{The performance of different levels of data augmentation methods on CIFAR-100. Compared to other approaches which focus on utilizing only one category, IPMix achieves state-of-the-art accuracy and robustness. }
\label{compare}
\vspace{-1.5em}
\end{wrapfigure}
IPMix integrates the three data augmentation techniques in a label-preserving fashion, effectively circumventing potential manifold intrusion and maintaining label consistency\cite{Sohn2020FixMatchSS}. Furthermore, inspired by prior work, IPMix eliminates the need to search for an optimal data augmentation policy, thus reducing computational costs. By addressing these challenges, IPMix has achieved significant improvements, as depicted in Figure \ref{compare}. In comparison to other methods that focus on leveraging one of these categories for enhancement, IPMix achieves state-of-the-art results in accuracy and robustness.

Since IPMix involves different levels of data augmentation techniques, it naturally motivates us to design a novel mixing method for better information fusion. Previous research has demonstrated that enhancing training data diversity~\cite{Zhang2017mixupBE,Devries2017ImprovedRO,Wang2021AugMaxAC} and image structural complexity~\cite{Lopes2020AffinityAD,Baradad2021LearningTS} is crucial for improving model robustness. The structural complexity of synthetic data, such as fractals and statistical information, can bolster model performance through pre-training~\cite{Kataoka2021PreTrainingWN} or blending with clean images~\cite{Hendrycks2021PixMixDP}. For better data integration, IPMix mixes clean images with synthetic pictures at different scales by random mixing to improve structural complexity, which can generate more diverse images to improve robustness.

Building on the enhancement of corruption robustness, we further extend IPMix's capabilities to enhance various safety metrics to fulfill the demands of constructing secure and reliable systems in real-world situations~\cite{Hendrycks2021UnsolvedPI}. We demonstrate that IPMix improves numerous safety metrics, including corruption robustness, calibrated uncertainty estimates, adversarial robustness, anomaly detection, and prediction consistency. On CIFAR-10-C and CIFAR-100-C, IPMix achieves the best results across different architectures. On ImageNet, IPMix outperforms previous methods and gains a substantial improvement on various safety measure benchmarks, achieving state-of-the-art or comparable results on ImageNet-R, ImageNet-A, and ImageNet-O~\cite{Hendrycks2020TheMF,Hendrycks2019NaturalAE}. 

 \begin{figure*}[t]
\vspace{-2em}
\begin{center}
\centerline{\includegraphics[width=1\columnwidth]{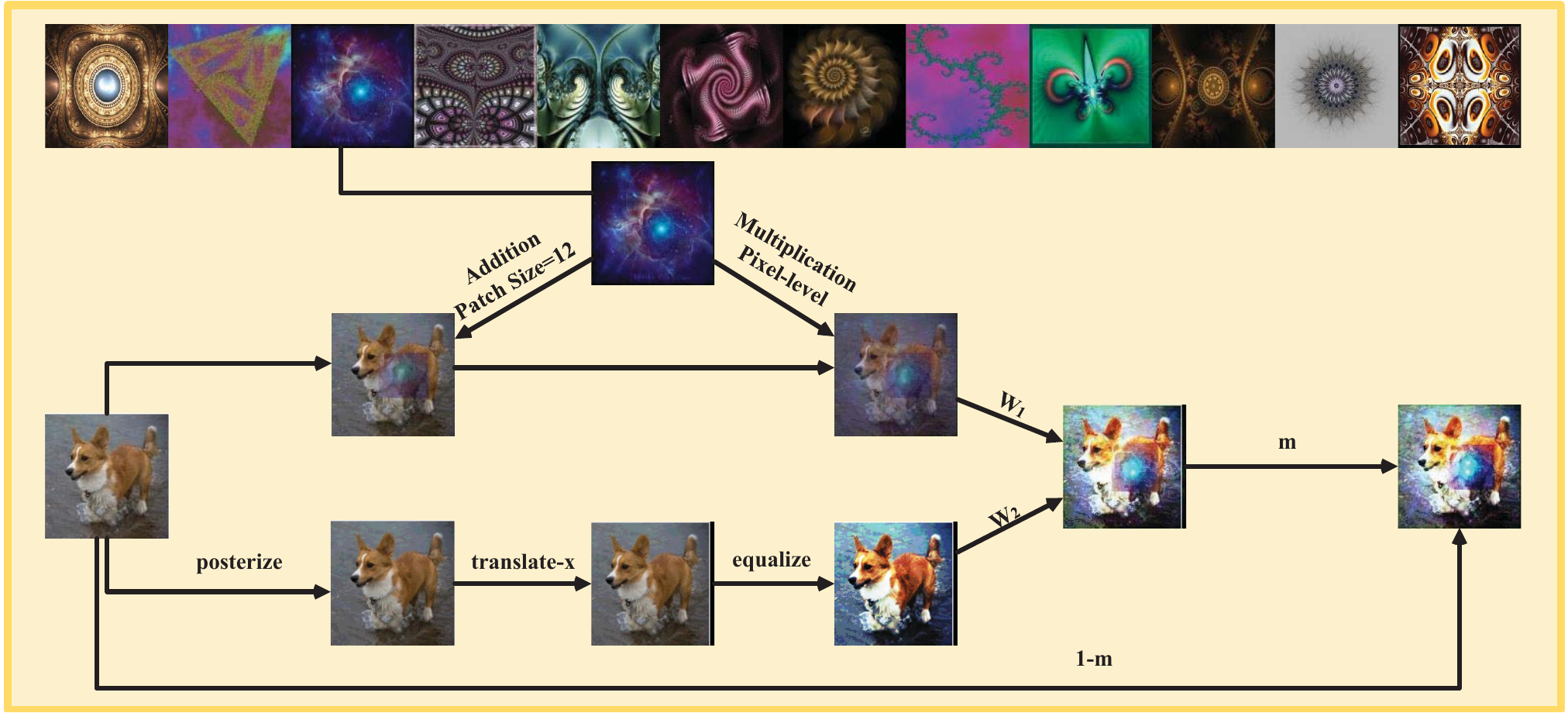}}
\caption{Top: Sample fractals from IPMix set. Bottom: An example of IPMix applied on a dog image, $k$ = 2, $t$ = 3. We randomly select P (pixel and patch) data augmentation methods and image-level data augmentation methods to generate a highly diverse set of augmented images. We sample $w_k$ ($k$ = 2, in this case) from Dirichlet distribution and use skip connection ($m$ sample from a Beta distribution) to maintain semantic consistency.}
\label{IPMix}
\end{center}
\vspace{-1.5em}
\end{figure*}

\section{Related Works}
\subsection{Data Augmentation}
Data augmentation is crucial to the success of modern neural networks, contributing significantly to the improvement of model generalization performance. The presented data augmentation approaches can be classified into three high-level categories: image-level, pixel-level, and patch-level augmentations.

\textbf{Image-level data augmentation.} 
Image-level data augmentation methods are commonly label-preserving, applying transformations on the whole image to improve data diversity. AutoAugment~\cite{Cubuk2019AutoAugmentLA} utilizes reinforcement learning to automatically search optimal compositions of transformations. Adversarial AutoAugment~\cite{Zhang2019AdversarialA} generates adversarial images to extend data and produces a dynamic policy during training. TrivialAugment~\cite{Mller2021TrivialAugmentTY} randomly selects an operation and the magnitude to reduce search space and improve performance. AugMix~\cite{Hendrycks2019AugMixAS} uses multiple transformations to create high diversity of augmented images, achieving state-of-the-art results on corruption robustness and calibration. AugMax~\cite{Wang2021AugMaxAC}  unifies diversity and hardness to search for the worst-case mixing strategy. PRIME~\cite{Modas2021PRIMEAF} uses max-entropy image transformations to boost model corruption robustness. 

\textbf{Pixel-level data augmentation.} 
Pixel-level data augmentation methods mix images using pixel-wise weighted averages. MixUp~\cite{Zhang2017mixupBE} generates augmented images by linearly interpolating between two randomly selected images and their corresponding labels. Manifold MixUp~\cite{Verma2018ManifoldMB} encourages neural networks to learn smooth interpolations between data points in the hidden layers, improving accuracy by comparison with MixUp. PixMix~\cite{Hendrycks2021PixMixDP} utilizes structural complexity synthetic pictures, such as fractals and feature visualizations, to improve model performance. Our work shared similarities with PixMix, but we use multi-scale information and better information fusion methods to train robust models by leveraging more diverse examples.

\textbf{Patch-level data augmentation.} 
Patch-level data augmentation methods mask or replace parts of the original image with different information. CutOut~\cite{Devries2017ImprovedRO} randomly masks out regions of a clean image to learn less discriminative portions, thereby improving accuracy. CutMix~\cite{Yun2019CutMixRS} replaces a patch of an original image with another randomly picked image to improve performance. Patch Gaussian~\cite{Lopes2019ImprovingRW}, which inputs a patch of Gaussian noise into the clean picture, combines the improved accuracy of CutOut with the noise robustness of Gaussian. SaliencyMix~\cite{Uddin2020SaliencyMixAS}, based on the maximum intensity pixel local in the saliency map, replaces a square patch of the original image with salient information from another image. TokenMix~\cite{Liu2022TokenMixRI} improves the performance of vision transformers by partitioning the mixing region into multiple separated parts and mixing two images at the token level. AutoMix~\cite{Liu2021AutoMixUT} optimizes both the mixed sample generation task and the mixup classification task in a momentum training pipeline with corresponding sub-networks in a bi-level optimization framework.

\subsection{Safety Measures}
When deploying network models in real-world scenarios, it is crucial to consider comprehensive security measures beyond standard accuracy. Implementing unsafe machine learning systems in high-stakes environments~\cite{Finlayson2018AdversarialAA, Papernot2018SoKSA, Skaf2020ApplyingNA} can lead to incalculable losses. With the rise of multimodal large language models (MLLMs)~\cite{Huang2023LanguageIN,OpenAI2023GPT4TR,Shen2023HuggingGPTSA}, safety issues are receiving increasing attention because their superior performance still makes mistakes. For example, GPT-4~\cite{OpenAI2023GPT4TR} may be confidently wrong in its predictions and disturbed by adversarial questions. Previous research has proposed various safety measures, including but not limited to robustness and calibration.

\textbf{Robustness.} Corruption robustness considers how to improve the model resistance to unseen natural perturbations under data distribution shifts. As a variant of the original ImageNet, ImageNet-C~\cite{Hendrycks2018BenchmarkingNN} consists of 15 diverse commonplace corruptions belonging to different categories with five levels of severity, regarded as a general benchmark for corruption robustness. In addition to natural corruption, Hendrycks et al.~\cite{Hendrycks2020TheMF} demonstrate that models should measure generalization to various abstract visual renditions. The robustness of adversarial attacks focuses on defending against imperceptible perturbations to images~\cite{Dong2017BoostingAA}. Prior works have proposed that there is a trade-off between the robustness of adversarial perturbations and clean image accuracy~\cite{Xie2019AdversarialEI,Xie2018FeatureDF}. ImageNet-O and ImageNet-A~\cite{Hendrycks2019NaturalAE}, widely regarded as benchmarks for evaluating image classifier performance under shifts in both input data and label distributions, are utilized for anomaly detection.

\textbf{Calibration.} Calibrated prediction confidences, which indicate whether a model’s output should be trusted, are valuable for classification models in real-world settings. Bayesian approaches~\cite{Guo2017OnCO} are widely used to deal with uncertainty estimation. Kuleshov et al.~\cite{Kuleshov2018AccurateUF} utilize recalibration methods to solve the miscalibration of credible intervals. Ovadia et al.~\cite{Ovadia2019CanYT} provide a benchmark for evaluating the accuracy and uncertainty of models under data distributional shifts.

%\textbf{Anomaly detection.} Since models should ideally know what they do not know, they will need to identify when an example is anomalous. PatchCore proposes improving the performance used for industrial anomaly detectors by maximizing nominal information. CutPaste shows that cutting parts of images and pasting them at a random location with different scales can increase accuracy in self-supervised learning. ImageNet-O and ImageNet-A are considered benchmarks to test image classifier performance when input data distribution and label distribution shifts.

\subsection{Training with Synthetic Data}
Previous works have proved that training with synthetic data can improve performance on real datasets. Debidatta et al.~\cite{Dwibedi2017CutPA} discover that combining synthetic annotated datasets with real data can significantly improve the performance of instance detection. Baradad et al.~\cite{Baradad2021LearningTS} generate synthetic data by utilizing various procedural noise models. In addition, they find that naturalism and diversity are two important properties for synthetic data to achieve comparable results with real datasets. Kataoka et al.~\cite{Kataoka2021FormuladrivenSL,Kataoka2022ReplacingLR} propose a suite of datasets generated by formula-driven supervised learning. %Yamada et al. show that pre-training on 3D fractal datasets improves accuracy in 3D object detection. Baek $\&$ Shim show that training GANs on privacy-free synthetic datasets can improve performance and transfer into a low-shot dataset. 

\section{An Attempt to Integrate Existing Approaches}
\label{section:3}

\begin{wraptable}{r}{8.5cm}
 \vspace{-1em}
\caption{The combination of different levels of data augmentation. M, C and A are abbreviations for MixUp, CutMix, and AugMix, respectively.}
 \centering
\begin{tabular}{cccc}
    \toprule
    & \multicolumn{1}{c}{Classification} & \multicolumn{1}{c}{Robustness} & \multicolumn{1}{c}{Calibration} \\
    &Error($\downarrow$) &mCE($\downarrow$) &RMS($\downarrow$)\\
    \midrule
    Vanilla  &21.3 &50  &14.6  \\ 
    +M   & 20.5 (-0.8) & 45.9(-4.1)& 10.5(-4.1)\\
    +M+C & 20.2 (-1.1)  & 46.1(-3.9)& 22.7(+8.1)\\
    +M+C+A & 23.4 (+2.1)  & 50.1(+0.1) & 25.6(+11) \\
   \bottomrule[1pt]
\end{tabular}
 \vspace{-1em}
\label{combination}
\end{wraptable}
Some prior studies~\cite{Hendrycks2021PixMixDP,Yun2019CutMixRS} have suggested that combining different data augmentation techniques with existing methods can improve accuracy on standard datasets. However, these works merely employed simple combinations without considering the compatibility between methods at different levels. Simultaneously, these studies chose the clean accuracy as the sole evaluation metric and have not taken the model's safety performance into account. In this section, we select MixUp~\cite{Zhang2017mixupBE}, CutMix~\cite{Yun2019CutMixRS}, and AugMix~\cite{Hendrycks2019AugMixAS} as representative data augmentation approaches for pixel-level, patch-level, and image-level, respectively, to conduct combination experiments of these approaches on CIFAR-100. Please refer to Appendix~\ref{app:combination}
for more details about the combination experiments.

Results on Table \ref{combination} demonstrate that simply combining different data augmentation methods may significantly impair model performance. This could be attributed to the excessive perturbation of training data caused by the combination of these methods, making the newly generated samples more challenging to identify and impacting the model's ability to learn useful features, leading to performance degradation. When multiple label-variant methods are combined, manifold intrusion issues may be more likely to arise. One possible solution for better information integration is to incorporate approaches (\eg, MixUp) into search-based data augmentation techniques~\cite{Cubuk2019RandaugmentPA,Mller2021TrivialAugmentTY}. However, searching the space for an optimal DA policy will bring expensive computation. Furthermore, this approach aims at improving clean accuracy and does not consider the overall safety performance.

\begin{figure}[t]
 \vspace{-1em}
\begin{center}
\centerline{\includegraphics[scale=0.168]{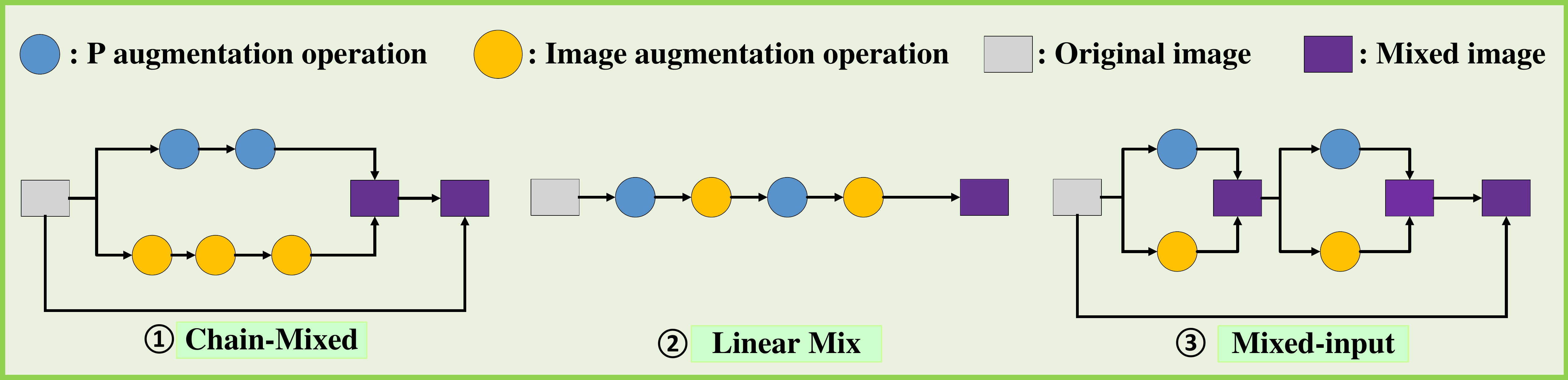}}
\caption{Different mixing framework of IPMix. P augmentation operation represents pixel-level and patch-level augmentation operations. \textcircled{\scriptsize{1}} Utilizing P operations and image-level operations in different chains and mixing the results. \textcircled{\scriptsize{2}} A clean image is randomly carried out by P operations or image-level operations in linear combinations to generate an IPMix image. \textcircled{\scriptsize{3}} leveraging the mixed image as a new input.}
\label{structure}
\end{center}
 \vspace{-1.5em}
\end{figure}

\section{IPMix: A Simple Method for Training Robust Classifiers}
In this section, we propose IPMix, which integrates three levels of data augmentation methods into a label-preserving approach, comprehensively improving safety metrics without sacrificing clean accuracy. We first demonstrate how to merge various techniques into a coherent framework and then propose novel approaches to achieve superior information fusion.

\subsection{Integrates Different Levels into A Coherent Approach}
\textbf{Pixel-level \& Patch-level.}  As a label-preserving data augmentation approach, IPMix uses the equation below to mix two input images:
\begin{equation}
    \tilde{x} = \mathit{B} \odot x_1 + (\mathit{I} - \mathit{B}) \odot x_2 \label{mix_x} \\
\end{equation}
Where $x_1$ is the input image and $x_2$ represents an unlabeled synthetic image (\eg, fractals, spectrum, or auto-generated contours). $\mathit{B}$ is a mask matrix suitable for both patch-level and pixel-level data augmentation methods, and $\mathit{I}$ is a binary mask filled with ones, having the same dimensions as $\mathit{B}$. $\odot$ represents the element-wise product. When performing mixing operations at the patch level, we choose a patch of random size and position from $\mathit{B}$, with a value of $\lambda$ (sample from Beta distribution) in this range and a value of 1 in other areas, which ensures that except for the mixing patch, the rest of the generated image comes from $x_1$. When performing mixing operations at the pixel level, we treat the entire image as a patch, with a value of $\lambda$. To make it efficient, we adopt fractals as representatives of synthetic data. However, IPMix is insensitive to mixing sets change, as shown in Table~\ref{source}.

Fractals are geometric shapes with structural complexities and natural geometries. While previous works~\cite{Kataoka2021PreTrainingWN,Anderson2021ImprovingFP} merely use iterated function systems (IFS) to create fractal data, we employ the Escape-time Algorithm for generating "orbit trap" complex fractals to enhance dataset complexity and diversity. Please refer to Appendix~\ref{app:fractal}
for details about generating fractal images.

The above-described method provides two key advantages: (1) We utilize a simple approach to combine operations of two levels, facilitating better information fusion. (2) Our method is label-preserving, ensuring it is not affected by manifold intrusion while eliminating the need for label smoothing~\cite{Mller2019WhenDL}. In the following sections, we refer to the method used in Eq. (\ref{mix_x}) as \textbf{P-level} data augmentation, signifying the employment of both patch-level and pixel-level methods.

\textbf{Image-level.} IPMix leverages various augmentation techniques and compositions to create a new image that does not deviate significantly from the original. Drawing inspiration from previous works~\cite{Mller2021TrivialAugmentTY,Hendrycks2019AugMixAS}, we randomly sample operations from PIL (\eg, brightness, sharpness) and randomly sample strengths to enhance the diversity of training data without expensive searching. Notably, these operations are disjoint from ImageNet-C corruptions, ensuring the robustness test's validity.

\begin{wraptable}{r}{8.8cm}
    \caption{Results are reported on CIFAR-100 and CIFAR-100-C with ResNeXt-29. The Chain-Mixed achieves the most balanced result on these metrics. Bold is best.}
    \vspace{2mm}
    \centering
    \begin{tabular}{cccc}
    \toprule
    & \multicolumn{1}{c}{Classification} & \multicolumn{1}{c}{Robustness} & \multicolumn{1}{c}{Calibration} \\
    &Error($\downarrow$) &mCE($\downarrow$) &RMS($\downarrow$)\\
    \midrule
    \textbf{Chain-Mixed}    &18.3 & 28.1 & 3.8   \\ 
    Linear Mix     &\textbf{18.2} & \textbf{27.4} & 13.5  \\
    Mixed Input   &19.8 & 29.6 & \textbf{3.6}  \\
   \bottomrule[1pt]
    \end{tabular}
     \vspace{-1em}
\label{framework}
\end{wraptable}
\textbf{The IPMix framework.} To determine the most effective methods for combining P-level and image-level, we conducted experiments using different mixing structures to generate a diverse set of IPMix images, as illustrated in the Figure \ref{structure} and Table \ref{framework}. While Linear Mix achieves excellent results in clean accuracy and corruption robustness, it performs poorly in calibrated prediction confidence. Mixed Input performs better in calibration but is inferior in accuracy and corruption robustness compared to Chain-Mixed. Consequently, we chose Chain-Mixed as the default framework for IPMix. Furthermore, the experimental results highlight the potential of establishing a general framework for integrating various data augmentation methods.
\subsection{Multi-scale Information Fusion}
IPMix can enhance the diversity and the structural complexity of training data to improve model performance. However, we found that simple mixing methods restrict the model's capabilities. 
\begin{wrapfigure}{r}{5cm}
 \vspace{-1em}
\centering
\includegraphics[width=0.35\textwidth]{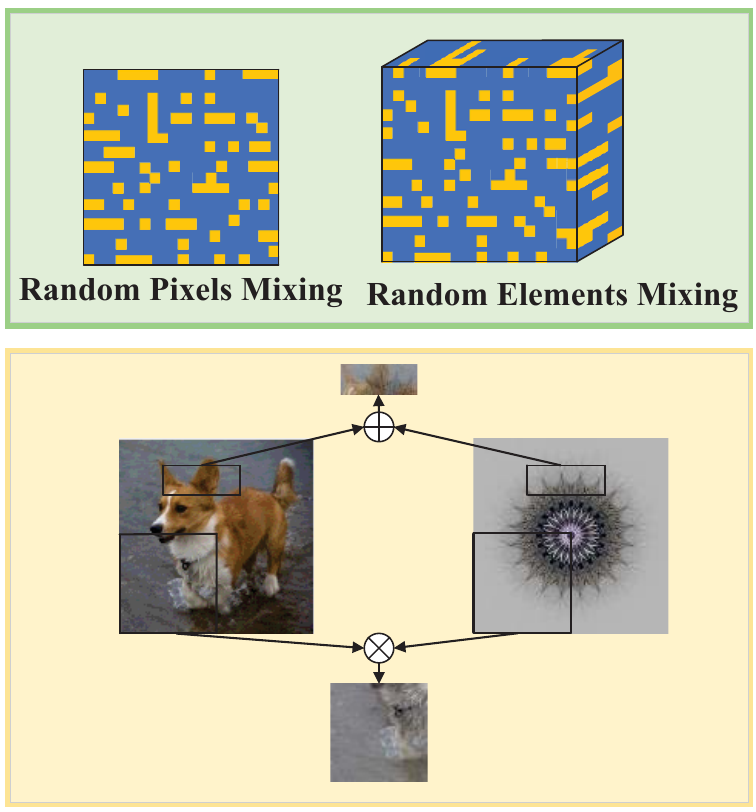}
\caption{Top: Examples of random mixing operations. Bottom: Examples of IPMix-Scar mixing and IPMix-Square mixing. }
\label{random_mixing}
 \vspace{-1em}
\end{wrapfigure}To overcome this issue, we use random mixing and scar-like image patches for achieving more effective information fusion.

\textbf{Random mixing.}
In previous data augmentation works, it is typical to either linearly mix two images or extract specific image features, such as saliency~\cite{Kim2021CoMixupSG,Uddin2020SaliencyMixAS}, which requires additional computations, for image mixing. As IPMix incorporates various levels of operations, its objective is to enhance the mixing of images, ultimately increasing data diversity. To accomplish this objective, IPMix employs four mixing operations: addition, multiplication, random pixel mixing, and random element mixing~\cite{Summers2018ImprovedMD}. Random pixel mixing creates a binary mask of size $H\times W \times1$ that operates on each channel sequentially, while random element mixing generates a binary mask of size $H\times W \times3$ (RGB) that applies to all channels simultaneously. An example is shown in Figure \ref{random_mixing}. The experiments in Appendix~\ref{app:ablation} show that both operations are beneficial to better information mixing between images and fractals. 
%~\ref{app:ablation} 

\textbf{Scar-like image patches.}
IPMix-Scar employs a long, thin rectangular box filled with an image patch to enhance dataset diversity, which has proven effective for anomaly detection~\cite{Li2021CutPasteSL}. An example of patch mixing is illustrated in Figure \ref{random_mixing}. First, IPMix randomly selects a point and a scar or square of the previously chosen size from the current image. Next, IPMix crops corresponding portions of the current image and the fractal picture and combine them. 

Finally, we obtain IPMix, which employs various levels of data augmentation to create diverse transformations with image structural complexity and data diversity. Figure \ref{IPMix} displays an example of IPMix, where \emph{\textbf{k}} denotes the number of augmented chains, and \emph{\textbf{t}} represents the maximum number of times an image can be augmented. The algorithm of IPMix is summarized in Appendix~\ref{app:algo_IPMix}.

\section{Experiments}
In this section, we showcase the significant performance improvements brought by IPMix on clean datasets in multiple settings. We present the evaluation results of IPMix for image classification on three datasets—CIFAR-10, CIFAR-100~\cite{Krizhevsky2009LearningML}, and ImageNet~\cite{Deng2009ImageNetAL}—across various models. Besides clean Classification, we assess IPMix on diverse safety tasks, including adversarial attack robustness, corruption robustness, prediction consistency, calibration, and anomaly detection. Please refer to Appendix~\ref{app:eva}
for details about the evaluation metrics. Lastly, we evaluate the properties of IPMix in thorough ablation studies and compare our approach with different levels of methods.

We evaluate IPMix on CIFAR-10-C, CIFAR-100-C, and ImageNet-C to measure its resistance to corruption data shifts. We test IPMix on CIFAR-10-P, CIFAR-100-P, and ImageNet-P to measure network prediction stability against minor perturbations. To thoroughly demonstrate our method's capabilities, we assess it on supplementary datasets, including ImageNet-R, ImageNet-O, and ImageNet-A. Experiments on these datasets validate our approach's robustness under real-world distribution shifts.

\begin{table}[t]
 \vspace{-0.5em}
\caption{Clean Error for IPMix on CIFAR-10 and CIFAR-100, lower is better. Top : CIFAR-10. Bottom : CIFAR-100. Mean and standard derivation over three random
seeds is shown for each experiment. Bold is best.}
\vspace{0.5em}
\centering\resizebox{\textwidth}{!}{
\begin{tabular}{cccccccc}
    \toprule
    & Vanilla & MixUp & CutOut & CutMix & AugMix & PixMix & IPMix  \\
    \midrule
    WideResNet40-4    &4.4$_{(\pm0.05)}$ & 3.8$_{(\pm0.06)}$ & \textbf{3.6}$_{(\pm0.05)}$ & 4.0$_{(\pm0.04)}$ & 4.3$_{(\pm0.08)}$ & 4.1$_{(\pm0.08)}$ & 4.0$_{(\pm0.06)}$\\ 
    WideResNet28-10   &3.8$_{(\pm0.07)}$ & 3.6$_{(\pm0.08)}$ & 3.4$_{(\pm0.06)}$ & 3.4$_{(\pm0.05)}$ & 3.4$_{(\pm0.07)}$ & 3.8$_{(\pm0.13)}$ & \textbf{3.3}$_{(\pm0.08)}$\\
    ResNeXt-29        &4.3$_{(\pm0.04)}$ & \textbf{3.8}$_{(\pm0.11)}$ & 4.2$_{(\pm0.08)}$ & \textbf{3.8}$_{(\pm0.02)}$ & 4.2$_{(\pm0.05)}$ & \textbf{3.8}$_{(\pm0.09)}$ & \textbf{3.8}$_{(\pm0.07)}$\\
    ResNet-18          &4.4$_{(\pm0.05)}$ & 4.2$_{(\pm0.04)}$ & 4.1$_{(\pm0.05)}$ & \textbf{4.0}$_{(\pm0.04)}$ & 4.5$_{(\pm0.03)}$ & 4.4$_{(\pm0.05)}$ & 4.2$_{(\pm0.07)}$\\
    \midrule
    Mean               &4.2 & 3.9 &  \textbf{3.8} &  \textbf{3.8} & 4.1 & 4.0 & \textbf{3.8}\\
   \midrule
    WideResNet40-4    &21.3$_{(\pm0.11)}$ & 20.5$_{(\pm0.13)}$ & 19.9$_{(\pm0.11)}$ & 20.3$_{(\pm0.15)}$ & 20.6$_{(\pm0.15)}$ & 20.4$_{(\pm0.17)}$ & \textbf{19.4}$_{(\pm0.14)}$\\ 
    WideResNet28-10  &19.0$_{(\pm0.13)}$ & 18.4$_{(\pm0.12)}$ & 18.8$_{(\pm0.15)}$ & 18.0$_{(\pm0.11)}$ & 19.4$_{(\pm0.11)}$ & 18.3$_{(\pm0.13)}$ & \textbf{17.4}$_{(\pm0.25)}$\\
    ResNeXt-29         &20.4$_{(\pm0.11)}$ & 20.3$_{(\pm0.12)}$ & 19.6$_{(\pm0.13)}$ & 19.5$_{(\pm0.13)}$ & 20.4$_{(\pm0.13)}$ & 20.1$_{(\pm0.11)}$ & \textbf{18.3}$_{(\pm0.22)}$\\
    ResNet-18          &23.7$_{(\pm0.09)}$ & 21.0$_{(\pm0.07)}$ & 22.0$_{(\pm0.11)}$ & \textbf{20.8}$_{(\pm0.12)}$ & 23.0$_{(\pm0.14)}$ & 21.6$_{(\pm0.15)}$ & 21.6$_{(\pm0.23)}$\\
   \midrule
    Mean               &21.1 & 20.0 & 20.1 & 19.7 & 20.8 & 20.1 & \textbf{19.2}\\
   \bottomrule
\end{tabular} }
\label{accuracy}
\end{table}

\begin{table}[t]
\caption{Mean Corruption Error (mCE) for IPMix across architectures on CIFAR-10-C and CIFAR-100-C, lower is better. Top : CIFAR-10-C. Bottom : CIFAR-100-C. Bold is best.}
\vspace{0.5em}
\centering\resizebox{\textwidth}{!}{
\begin{tabular}{cccccccc}
    \toprule
    & Vanilla & MixUp & CutOut & CutMix& AugMix& PixMix & IPMix  \\
    \midrule % Replace \hline with \midrule
    WideResNet40-4    &26.4$_{(\pm0.14)}$ & 21$_{(\pm0.15)}$ & 25.9$_{(\pm0.13)}$ & 26$_{(\pm0.13)}$ & 10$_{(\pm0.12)}$ & 9.5$_{(\pm0.14)}$ & \textbf{8.6}$_{(\pm0.14)}$\\
    WideResNet28-10   &24.2$_{(\pm0.15)}$ & 19.2$_{(\pm0.17)}$ & 23.5$_{(\pm0.17)}$ & 25.1$_{(\pm0.13)}$ & 9.1$_{(\pm0.14)}$ & 8.7$_{(\pm0.14)}$ & \textbf{7.5}$_{(\pm0.17)}$\\
    ResNeXt-29         &27.5$_{(\pm0.11)}$ & 23.6$_{(\pm0.18)}$ & 27.3$_{(\pm0.18)}$ & 28.5$_{(\pm0.18)}$ & 11.3$_{(\pm0.15)}$ & 9.2$_{(\pm0.12)}$ & \textbf{8.6}$_{(\pm0.19)}$\\
    ResNet-18          &25$_{(\pm0.09)}$ & 20$_{(\pm0.15)}$ & 24.1$_{(\pm0.13)}$ & 24.7$_{(\pm0.19)}$ & 10.4$_{(\pm0.13)}$ & 9$_{(\pm0.11)}$ & \textbf{8.4}$_{(\pm0.17)}$\\
    \midrule % Replace \hline with \midrule
    Mean               &25.8 & 20.9 & 25.2 & 26 & 10 & 9.1 & \textbf{8.2}\\
    \midrule % Replace \hline with \midrule
    WideResNet40-4    &50$_{(\pm0.15)}$ & 45.9$_{(\pm0.19)}$ & 51.5$_{(\pm0.17)}$ & 50$_{(\pm0.19)}$ & 33.3$_{(\pm0.22)}$ & 31.1$_{(\pm0.19)}$ & \textbf{28.6}$_{(\pm0.15)}$\\
    WideResNet28-10  &48.5$_{(\pm0.21)}$ & 44.2$_{(\pm0.18)}$ & 48.2$_{(\pm0.15)}$ & 48.6$_{(\pm0.21)}$ & 31.5$_{(\pm0.21)}$ & 28.3$_{(\pm0.21)}$ &  \textbf{26.6}$_{(\pm0.29)}$\\
    ResNeXt-29         &51.4$_{(\pm0.19)}$ & 47.9$_{(\pm0.21)}$ & 51$_{(\pm0.17)}$ & 52.4$_{(\pm0.22)}$ & 34.1$_{(\pm0.24)}$ & 30.6$_{(\pm0.23)}$ & \textbf{28.1}$_{(\pm0.31)}$\\
    ResNet-18          &50$_{(\pm0.18)}$ & 45.5$_{(\pm0.21)}$ & 50.2$_{(\pm0.19)}$ & 50.8$_{(\pm0.24)}$ & 35$_{(\pm0.25)}$ & 31.4$_{(\pm0.21)}$ & \textbf{29.9}$_{(\pm0.29)}$\\
    \midrule % Replace \hline with \midrule
    Mean               &50 & 45.9 & 50.2 & 50.5 & 33.4 & 30.3 & \textbf{28.3}\\
   \bottomrule
   \label{robustness}
   \end{tabular}}
    \vspace{-1.5em}
\end{table}

\subsection{Evaluation on CIFAR}
We experiment with different backbone architectures on CIFAR-10 and CIFAR-100, including 40-4 Wide ResNet~\cite{Zagoruyko2016WideRN}, 28-10 Wide ResNet, ResNeXt-29~\cite{Xie2016AggregatedRT}, and Resnet-18~\cite{He2015DeepRL}. We compare IPMix with various data augmentation methods, including CutOut, MixUp, CutMix, AugMix, and PixMix. Please refer to Appendix~\ref{app:exp_config} for more details about the training configurations.
%~\ref{app:exp_config}

\textbf{Accuracy.}
In Table \ref{accuracy}, we demonstrate that IPMix improves standard accuracy across architectures. In comparison with other approaches, IPMix achieves the best or comparable accuracy, showing the improvement of safety measures is not at the cost of hurting clean accuracy.

\textbf{Corruption robustness.}
Results show that IPMix substantially improves corruption robustness across architectures. Compared to AugMix on CIFAR-100-C, IPMix achieves \textbf{4.7\%}(40-4) and \textbf{4.9\%}(28-10) improvement on WideResNet, \textbf{6\%} on ResNeXt, and \textbf{5.1\%} on ResNet. Table \ref{robustness} demonstrates that IPMix achieves state-of-the-art results on both CIFAR-10-C and CIFAR-100-C.

\begin{figure}[t]
 \vspace{-2em}
\begin{center}
\centerline{\includegraphics[scale=0.67]{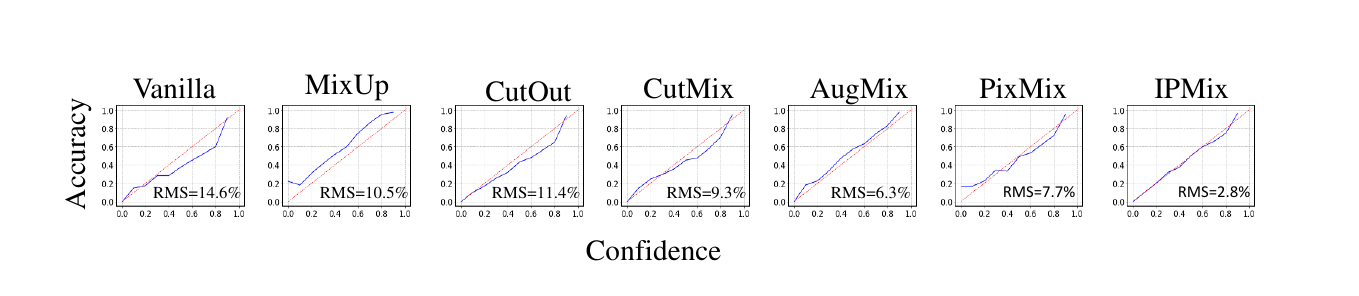}}
 \vspace{-0.5em}
\caption{The results of calibration on CIFAR-100. IPMix achieves the lowest RMS error in all data augmentation methods, improving \textbf{11.8\%} by comparing with Vanilla.}
\label{calibration}
\end{center}
 \vspace{-1.5em}
\end{figure}

\textbf{Calibration.}
We utilize RMS calibration error~\cite{Hendrycks2018DeepAD} to evaluate the empirical frequency of correctness.  As depicted in Figure \ref{calibration}, IPMix surpasses other methods, achieving state-of-the-art results.

\begin{figure}[t]
\begin{center}
\centerline{\includegraphics[scale=0.41]{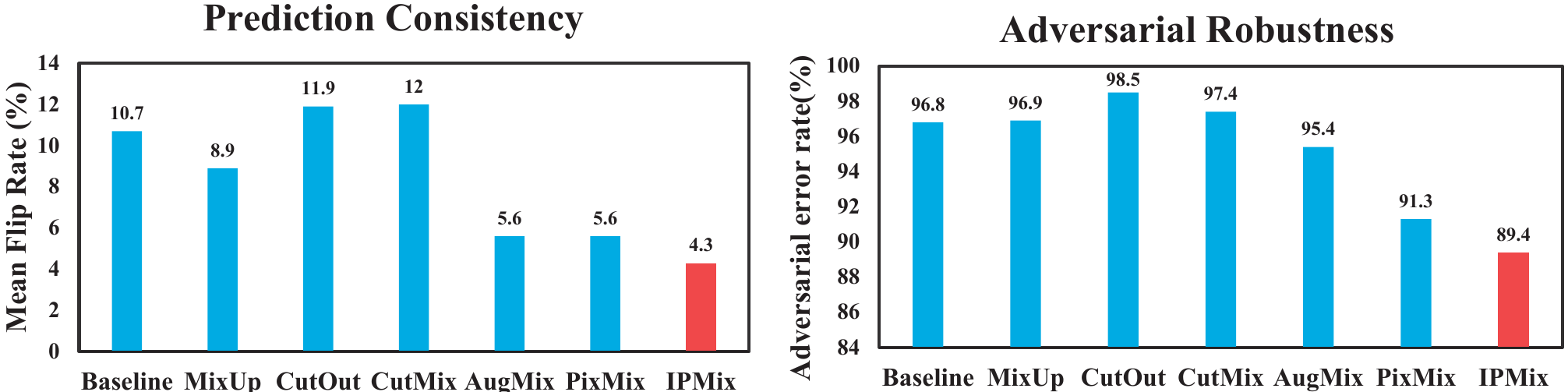}}
\caption{\textbf{Left: prediction consistency.} \textbf{Right: adversarial robustness.} IPMix achieves the best results on both metrics, demonstrating its ability to improve overall security performance.}
\label{p+adv}
\end{center}
 \vspace{-2em}
\end{figure}

\textbf{Prediction consistency.}
We leverage the mean flip rate (mFR) to evaluate prediction consistency on CIFAR-10-P and CIFAR-100-P~\cite{Hendrycks2018BenchmarkingNN}. IPMix achieves the lowest mFR, as shown in Figure \ref{p+adv}.

\textbf{Adversarial robustness.}
This measure evaluates the resistance of adversarially perturbed by projected gradient descent. We utilize PGD~\cite{Madry2017TowardsDL} to verify the adversarial robustness of image classifiers. The results in Figure \ref{p+adv} show that IPMix achieves the lowest error.

\begin{table*}[ht]
    \caption{The results of IPMix on ImageNet. For Anomaly Detection, we test the accuracy on ImageNet-A and AUPR on ImageNet-O, higher is better. IPMix achieves round improvement over various data augmentation methods. Bold is best, and underline is second. }
    \vspace{0.5em}
    \renewcommand{\arraystretch}{1.1}
    \centering\resizebox{\textwidth}{!}{
    \begin{tabular}{ccccccccccc}
        \Xhline{1pt}
         & \multicolumn{1}{c}{Classification} & \multicolumn{2}{c}{Robustness} & \multicolumn{1}{c}{Consistency} &\multicolumn{3}{c}{Calibration} &\multicolumn{2}{c}{Anomaly Detection}\\
         \cmidrule(r){2-2}
         \cmidrule(r){3-4}
         \cmidrule(r){5-5}
         \cmidrule(r){6-8}
         \cmidrule(r){9-10}
         &Clean&ImageNet-C&ImageNet-R& ImageNet-P & C & R & A & ImageNet-A & ImageNet-O\\
         &Error($\downarrow$)&mCE($\downarrow$)&Error($\downarrow$)& mFR($\downarrow$) & RMS($\downarrow$) & RMS($\downarrow$) & RMS($\downarrow$) & Classification($\uparrow$) & AUPR($\uparrow$)\\
        \midrule
        Vanilla & 23.9 & 78.6 & 64 & 57.7 & 12 & 19.9 & 47 & 2.2 & 16.2\\
        MixUp\cite{Zhang2017mixupBE} & 22.7 & 76.5 & 62.4 & 54.6 & 9.3 & 41.7 & 49.3 & 5.2 & 16.1\\
        CutOut\cite{Devries2017ImprovedRO} & 22.6 & 73.1 & 64.6 & 57.9 & 11.3 & 19.7 & 46.3 & 4.7 & 15.9\\
        CutMix\cite{Yun2019CutMixRS} & 22.9 & 77.2 & 66.5 & 58.1 & 9.6 & 44.2 &48 & \textbf{7.2} & 16.5\\
        AugMix\cite{Hendrycks2019AugMixAS} & 22.6 & 68.5 & 61.8 & 52.3 & 8.1 & 13.1 & \underline{43.5} & 3.8 & \underline{17.4}\\
        AugMax\cite{Wang2021AugMaxAC} & 22.9 & 67.4 & 62.1 & 54.6 & 8.8 & \underline{12.1} & 44.7 & 3.9 & 17.1\\
        PixMix\cite{Hendrycks2021PixMixDP} & \underline{22.4} & \underline{65.4} & \underline{59.8} & \underline{50.8} & \underline{7.2} & 12.3 & 44 & 5.9 & 17.3\\
        IPMix & \textbf{22.2} & \textbf{63} & \textbf{57.4} & \textbf{48.5} & \textbf{7.1} & \textbf{7} & \textbf{30} & \underline{6.6} & \textbf{18.2}\\

        \Xhline{1pt}
    \end{tabular} }
    \vspace{-0.5em}
    \label{imagenet}
\end{table*}
\subsection{Evaluation on ImageNet}
For ImageNet experiments, we compare different data augmentation methods, including MixUp, CutOut, CutMix, AugMix, AugMax~\cite{Wang2021AugMaxAC}, and PixMix. We utilize SGD optimizer with an initial learning rate of 0.01 to train ResNet-50 for 180 epochs following a cosine decay schedule. Please refer to Appendix~\ref{app:exp_config} for more details about the training configurations.
%~\ref{app:exp_config}

IPMix achieves state-of-the-art or comparable performances on a broad range of safety measures, as shown in Table \ref{imagenet}. Compared with other methods, IPMix improves the resistance of out-of-distribution shifts without reducing clean accuracy. On corruption robustness, IPMix outperforms Vanilla by \textbf{15.6\%} and  AugMix by \textbf{5.5\%}, achieving state-of-the-art results. On ImageNet-R, IPMix demonstrates the ability to improve rendition robustness, increasing by \textbf{6.6\%} by comparison with Vanilla. On ImageNet-P, IPMix improves mFR by \textbf{9.2\%} over Vanilla and \textbf{2.3\%} over PixMix. On calibration tests, IPMix surpasses all methods on ImageNet-C, ImageNet-R, and ImageNet-A, improving RMS by \textbf{0.1\%}, \textbf{5.1\%}, and \textbf{13.5\%} by comparison with the second-best approach. Furthermore, IPMix achieves convincing results on ImageNet-A and ImageNet-O, demonstrating its exceptional ability in anomaly detection. The results demonstrate that IPMix can roundly improve safety metrics.

\begin{table*}[ht]
    \caption{Ablation results of different components of IPMix on ImageNet with ResNet-50. }
    \renewcommand{\arraystretch}{1.1}
    \centering\resizebox{\textwidth}{!}{
    \begin{tabular}{ccccccccccc}
        \Xhline{1pt}
         & \multicolumn{1}{c}{Classification} & \multicolumn{2}{c}{Robustness} & \multicolumn{1}{c}{Consistency} &\multicolumn{3}{c}{Calibration} &\multicolumn{2}{c}{Anomaly Detection}\\
         \cmidrule(r){2-2}
         \cmidrule(r){3-4}
         \cmidrule(r){5-5}
         \cmidrule(r){6-8}
         \cmidrule(r){9-10}
         &Clean&ImageNet-C&ImageNet-R& ImageNet-P & C & R & A & ImageNet-A & ImageNet-O\\
         &Error($\downarrow$)&mCE($\downarrow$)&Error($\downarrow$)& mFR($\downarrow$) & RMS($\downarrow$) & RMS($\downarrow$) & RMS($\downarrow$) & Classification($\uparrow$) & AUPR($\uparrow$)\\
        \midrule
        IPMix & \textbf{22.2} & \textbf{63} & \textbf{57.4} & \textbf{48.5} & \textbf{7.1} & \textbf{7} & \textbf{30} & \textbf{6.6} & \textbf{18.2}\\
        w/o patch & 22.8$_{(\pm0.11)}$ & 65.1$_{(\pm0.16)}$ & 58.8$_{(\pm0.11)}$ & 49.1$_{(\pm0.15)}$ & 7.8$_{(\pm0.11)}$ & 7.4$_{(\pm0.08)}$ & 31.1$_{(\pm0.01)}$ & 6$_{(\pm0.01)}$ & 17.7$_{(\pm0.02)}$\\
        w/o pixel & 23.1$_{(\pm0.15)}$ & 65.6$_{(\pm0.19)}$ & 59.3$_{(\pm0.13)}$ & 49.5$_{(\pm0.17)}$& 8.2$_{(\pm0.13)}$ & 7.4$_{(\pm0.09)}$ & 32.4$_{(\pm0.11)}$ & 5.6$_{(\pm0.01)}$ & 17.2$_{(\pm0.03)}$\\
        w/o image & 23.5$_{(\pm0.16)}$ & 66.2$_{(\pm0.21)}$ & 59.5$_{(\pm0.17)}$ & 49.6$_{(\pm0.14)}$ & 8.8$_{(\pm0.13)}$ & 8.1$_{(\pm0.13)}$ & 33.5$_{(\pm0.13)}$ & 6.5$_{(\pm0.02)}$ & 17.8$_{(\pm0.03)}$\\
        \Xhline{1pt}
    \end{tabular} }
    \vspace{-1em}
    \label{imagenet_abolation}
\end{table*}

\subsection{Ablation Study}
\begin{wraptable}{r}{8.5cm}
\vspace{-1em}
        \caption{Ablation results of different components of IPMix on CIFAR-100. Mean and standard derivation over three random seeds is shown for each experiment. Bold is the best.}
    \centering
    \begin{tabular}{cccc}
        \toprule
            & Classification & Robustness & Calibration \\
        \midrule
        IPMix &\textbf{19.4} &\textbf{28.6} &\textbf{2.8}   \\ 
        \midrule
        w/o patch  &19.7$_{(\pm0.13)}$ &30 $_{(\pm0.21)}$ &4.6 $_{(\pm0.07)}$  \\
        w/o pixel  &19.6 $_{(\pm0.09)}$&33 $_{(\pm0.35)}$ &8.2 $_{(\pm0.12)}$  \\
        w/o image  &20.1 $_{(\pm0.27)}$&34 $_{(\pm0.65)}$&8.6 $_{(\pm0.21)}$ \\
       \bottomrule[1pt]
    \end{tabular}
    \vspace{-1em}
    \label{parts}
\end{wraptable}
In this paragraph, we evaluate the properties of our approach by ablation experiments. We first study the influence of different parts of IPMix on performance and then assess the stability of IPMix under various mixing sources. Please refer to more ablation experiments in Appendix~\ref{app:ablation}. 

\textbf{Components of IPMix.} 
In this section, we evaluate the influence of different IPMix components on performance. We execute ablation experiments on the three primary IPMix constituents: image-level, patch-level, and pixel-level. The results show the indispensable contribution of each component to enhancing model performance, demonstrating that these approaches are complementary and that a unification among them is necessary to achieve robustness. The ablation experiment results are shown in Table \ref{imagenet_abolation} and Table \ref{parts}. Please refer to thorough analysis in Appendix~\ref{app:analysis}.

\begin{wraptable}{r}{8.5cm}
\vspace{-1em}
\caption{Ablation results on IPMix across different mixing sets.The results show that IPMix is insensitive to mixing sets change. }
\vspace{0.5em}
\centering
\begin{tabular}{cccc}
    \toprule
    \multirow{2}{*}{Mixing sets}& \multicolumn{1}{c}{Classification} & \multicolumn{1}{c}{Robustness} & \multicolumn{1}{c}{Calibration} \\
    &Error($\downarrow$) &mCE($\downarrow$) &RMS($\downarrow$)\\
    \midrule
    Fractal + FVis   &\textbf{19.4} & 28.8 & 3.3   \\ 
    FractalDB   &20 & 29 & 5.4  \\
    RCDB   &19.5 & \textbf{28.4} & 3.2  \\
    Dead Leaves   &\textbf{19.4} & 29.1 & 3.1  \\
    Spectrum   &19.8 & 29.2 & 4  \\
    fractals(ours) &\textbf{19.4} & 28.6 & \textbf{2.8}  \\
   \bottomrule[1pt]
\end{tabular}
\vspace{-1em}
\label{source}
\end{wraptable}
\textbf{Mixing sources.}
The excellent performance of IPMix is partly due to the structural complexity of fractal pictures. In this part, we examine the sensitivity of IPMix to different fractal sources on CIFAR-100. We report clean accuracy, corruption robustness, and calibration from different sources with WRN40-4.  Fractal + FVis is the default setting of PixMix, which consists of fractals and feature visualization. FractalDB~\cite{Kataoka2021FormuladrivenSL} consists of fractal images generated by Iterated Function System (IFS). RCDB~\cite{Kataoka2022ReplacingLR} consists of auto-generated contours. Dead Leaves and Spectrum generated from generative image models~\cite{Baradad2021LearningTS}. The full results show in Table \ref{source}.

\subsection{The Comparison with Different Levels of Method}
In this section, we perform an extensive performance comparison between IPMix and a range of existing methods using multiple metrics. We consider AutoAugment, RandAugment, and TrivialAugment~\cite{Mller2021TrivialAugmentTY} as representative image-level techniques, while SaliencyMix, PuzzleMix~\cite{Kim2020PuzzleME}, and Co-Mixup~\cite{Kim2021CoMixupSG} serve as typical patch-level techniques. For pixel-level methods, Manifold Mixup stands as our representative choice. IPMix does not require searching for the optimal DA policy like image-level techniques. In contrast to patch-level approaches, IPMix eliminates the need for saliency computations. The results in Table \ref{comparation} show that IPMix outperformed all other methods on all metrics.

\begin{figure}[t]
 \vspace{-2.5em}
\begin{center}
\centerline{\includegraphics[scale=0.4]{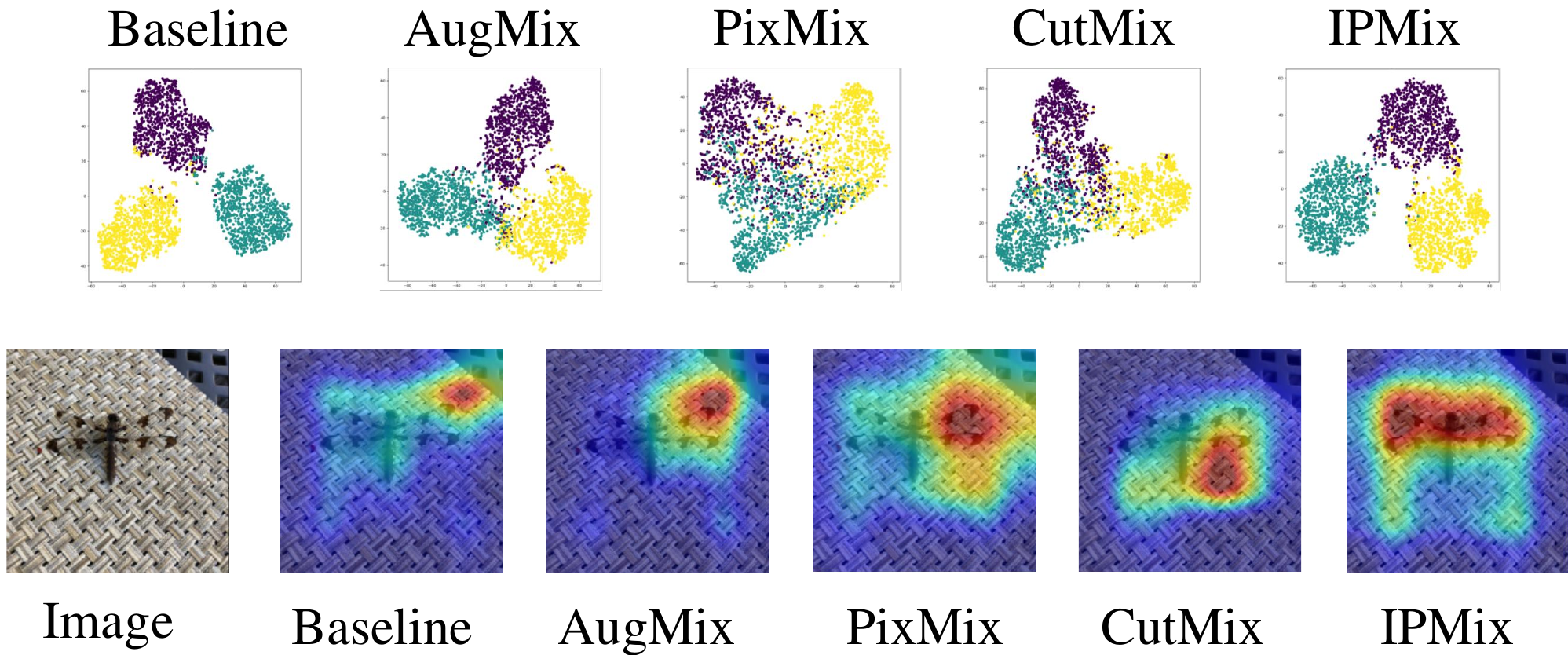}}
\caption{\textbf{Top: t-SNE visualization.} The features are from the penultimate layer of a WRN40-4 trained on CIFAR10. Compared with other approaches, IPMix has distinct boundaries between different category clusters and generates diverse samples to cover boundary areas, thereby improving the generalization ability. \textbf{Bottom: The Grad-CAM visualization}, with input images sourced from ImageNet-A, demonstrates that IPMix excels in identifying objects within complex scenarios.}
\label{tsne+cam}
\end{center}
 \vspace{-2em}
\end{figure}

\begin{table}[ht]
\vspace{-0.5em}
    \caption{Results of different augmentation methods on CIFAR-100 and CIFAR-100-${\mbox{C}}$ with 28-10 Wide ResNet. Bold is best.}
\centering
\vspace{0.5em}
\begin{tabular}{cccccc}
    \toprule
    \multirow{2}{*}{Methods}& \multicolumn{1}{c}{Classification}& \multicolumn{1}{c}{Robustness} & \multicolumn{1}{c}{Adversaries} &\multicolumn{1}{c}{Consistency} & \multicolumn{1}{c}{Calibration} \\
    &Error($\downarrow$) &mCE($\downarrow$) &Error($\downarrow$) &mFR($\downarrow$) &RMS($\downarrow$)\\
    \midrule
    AutoAugment~\cite{Cubuk2019AutoAugmentLA}  &17.7$_{(\pm0.11)}$ &38.4$_{(\pm0.15)}$ &97.8$_{(\pm0.22)}$  &8$_{(\pm0.06)}$  &7.9$_{(\pm0.06)}$  \\ 
    RandAugment~\cite{Cubuk2019RandaugmentPA}  &17.8$_{(\pm0.14)}$ &41.5$_{(\pm0.13)}$ &96.6$_{(\pm0.25)}$ &8.6$_{(\pm0.10)}$  &7.9$_{(\pm0.04)}$\\ 
    TrivialAugment~\cite{Mller2021TrivialAugmentTY} &17.9$_{(\pm0.13)}$ &96.3$_{(\pm0.21)}$  &35.4$_{(\pm0.23)}$ &7.3$_{(\pm0.07)}$ &8.7$_{(\pm0.04)}$  \\ 
    SaliencyMix~\cite{Uddin2020SaliencyMixAS}  &18.3$_{(\pm0.14)}$ &38.3$_{(\pm0.24)}$ &96.7$_{(\pm0.21)}$ &10.8$_{(\pm0.07)}$   &7.1$_{(\pm0.07)}$ \\ 
    PuzzleMix~\cite{Kim2020PuzzleME}  &18.1$_{(\pm0.11)}$ &37.9$_{(\pm0.21)}$ &96.1$_{(\pm0.23)}$ &10.5$_{(\pm0.04)}$ &7.5$_{(\pm0.08)}$\\
    Co-Mixup~\cite{Kim2021CoMixupSG}  &18.0$_{(\pm0.19)}$ &35.6$_{(\pm0.25)}$ &95.6$_{(\pm0.21)}$ &10.1$_{(\pm0.05)}$ &7.7$_{(\pm0.04)}$\\
    Manifold Mixup~\cite{Verma2018ManifoldMB} &18.8$_{(\pm0.21)}$ &51.3$_{(\pm0.23)}$ &93.4$_{(\pm0.17)}$ &29.9$_{(\pm0.28)}$ &10.2$_{(\pm0.09)}$\\
    IPMix  &\textbf{17.4$_{(\pm0.25)}$} &\textbf{26.6$_{(\pm0.29)}$} &\textbf{91.3$_{(\pm0.21)}$} &\textbf{4.2$_{(\pm0.11)}$} &\textbf{6.4$_{(\pm0.07)}$}\\
   \bottomrule[1pt]
\end{tabular}
 \vspace{-1em}
\label{comparation}
\end{table}

\section{Analysis of IPMix}
IPMix combines three levels of data augmentation into a unified, label-preserving technique to improve model performance. We believe that IPMix's superior performance is due to the increased data diversity and enhanced regularization effect. For a more intuitive demonstration of these effects, we utilize t-SNE and Class Activation Mapping (CAM)~\cite{Selvaraju2016GradCAMVE} for visualizations, as shown in Figure \ref{tsne+cam}.

\textbf{Increasing diversity.} IPMix increases the diversity of training data by mixing data at multiple levels, enabling the model to learn a greater variety of feature combinations and patterns. Furthermore, the integration of synthetic data from distinct distributions (\eg, fractals), further amplifies this diversity.

\textbf{Enhanced regularization effect.} The approach of mixing data also serves as a potent regularization technique. By randomly mixing samples, the model is compelled to learn more robust features rather than overly relying on specific sample or class characteristics, which reduces the risk of overfitting and enhances the model's performance in different environments.

\section{Conclusion}
We propose IPMix, which leverages different levels of augmentation techniques and image structural complexity to improve model performance. By employing random mixing methods, we facilitate more effective information fusion. The experimental results indicate that IPMix can significantly improve various safety metrics. We hope our work will attract attention to joining different methods into coherent and synergetic approaches to improve robustness and other safety measures. This adaptation is crucial given the growing importance of safety requirements in systems design.

\section*{Acknowledgement}
\vspace{-0.5em}
This work was under the help of Xi Yang. We thank him for his selfless help and valuable suggestions.

{
\small
\bibliographystyle{unsrt}
\bibliography{IPMix}
}

%%%%%%%%%%%%%%%%%%%%%%%%%%%%%%%%%%Appendix%%%%%%%%%%%%%%%%%%%%%%%%%%
\clearpage
\appendix

\textbf{Contents of the Appendices:}

Section~\ref{app:exp_config}. Details of experimental settings, hyperparameters, and configurations used in this paper.

Section~\ref{app:experiments}. Additional experiments of IPMix, including more ablation experiments, comparisons with other methods, and additional robustness experiments.

Section~\ref{app:eva}. The details of evaluation metrics.

Section~\ref{app:algo_IPMix}. The algorithm of IPMix.

Section~\ref{app:fractal}. The details about generating fractal images.

Section~\ref{app:combination}. The details about combination experiments.

Section~\ref{app:time}. Training time of IPMix.

Section~\ref{app:full_res}. Full results of IPMix across architecturess.

Section~\ref{app:CAM}. More CAM visualizations of IPMix.

Section~\ref{app:analysis}. The Analysis of Ablation Experiments. 

Section~\ref{app:Vit}. The Experiment Results on Transformer Architecture.

Section~\ref{app:drawbacks}. The Drawbacks of Different Levels of Methods.

Section~\ref{app:limit}. Limitation and Broader Impact.

\section{Experimental Settings}
\label{app:exp_config}
\subsection{CIFAR}
In this section, we share the training settings of IPMix on CIFAR. We experiment with various backbone architectures on CIFAR-10 and CIFAR-100, including 40-4 Wide ResNet, 28-10 Wide ResNet~\cite{Zagoruyko2016WideRN}, ResNeXt-29~\cite{Xie2016AggregatedRT}, and Resnet-18~\cite{He2015DeepRL}. We train ResNet and RexNeXt for 200 epochs, and all Wide ResNets for 100 epochs. We employ the SGD optimizer with a weight decay of 0.0001 and a momentum of 0.9. We randomly crop training images to 32$\times$32 resolution with zero padding and flip them horizontally. We compare IPMix with various data augmentation methods, including CutOut, MixUp, CutMix, AugMix, and PixMix. We select a CutOut size of 16$\times$16 pixels on CIAFR-10, and 8$\times$8 on CIFAR-100. For CutMix, we set CutMix probability as 0.5 and $\alpha$ = 1.0. We set $k$ = 3 in AugMix, and $k$ = 3, $\beta$ = 4 in PixMix for the best results. For IPMix, we set $k$ = 3, $t$ = 3, and randomly select patch sizes from 4, 8, 16, and 32 (pixel-level).  All experiments are conducted on a server with two NVIDIA GeForce RTX 3090 GPUs.
\subsection{ImageNet-1K}
For ImageNet experiments, we compare different data augmentation methods, including MixUp, CutOut, CutMix, AugMix, AugMax~\cite{Wang2021AugMaxAC}, and PixMix. Since regularization methods may require a greater number of training epochs to converge, we fine-tune a pre-trained ResNet-50 for 180 epochs. We utilize SGD optimizer with an initial learning rate of 0.01 following a cosine decay schedule, with a batch size of 256. For all approaches, we randomly crop training images to 224$\times$224 resolution with zero padding and flip them horizontally. We adopt $\alpha$ = 0.2 for MixUp and CutMix and select a CutOut size of 56$\times$56 pixels. For IPMix, we use $k$ = 3, $t$ = 3, and randomly select patch sizes from 4, 8, 16, 32, 64, and 256 (pixel-level). We set ${\lambda}$ = 12 and $n$ = 5 for AugMax-DuBIN, the same as the paper. 

\newpage

\section{Additional Experiments of IPMix }
\label{app:experiments}

\subsection{Ablation Exmperiments}
\label{app:ablation}
\textbf{IPMix hyperparameters.}
In this paragraph, we evaluate the hyperparameters sensitivity of IPMix. We examine two hyperparameters: the number of augmented chains $k$ and the maximum image augmentation times $t$ with clean accuracy and robustness. The results in Table \ref{app:hyper} demonstrate that IPMix is not sensitive to hyperparameters, showing the performance of IPMix is stable under change.

\textbf{Mixing operations ablation.}
In this paragraph, we test IPMix's mixing operation sensitivity. IPMix utilizes four different operations to improve model performance, including addition, multiplication, random pixels mixing, and random elements mixing. The results show in Table \ref{app:operations}.

\textbf{Patch mixing ablation.}
In this paragraph, we verify IPMix's patch variants, which can be divided into two categories, IPMix-Scar and IPMix-Square. The results in Table \ref{app:variants} show that PachtMix-Scar can improve model robustness.

\begin{table}[th]
\vspace{-1em}
\caption{We evaluate clean accuracy on CIFAR-100 and Mean Corruption Error (mCE) on CIFAR-100-C with WRN40-4. The performance of IPMix is not strongly associated with hyperparmeters.}
\vspace{0.5em}
\centering
\begin{tabular}{c|c|c|c}
    \toprule
    & $k$ = 2 & $k$ = 3 & $k$ = 4 \\
    \midrule
    $t$ = 2 &\makecell{19.5 \\ 29} &\makecell{19.3 \\ 28.9} &\makecell{19.3 \\ 29}   \\ 
    \midrule
    $t$ = 3    &\makecell{19.7 \\ \textbf{28.5}} &\makecell{\textbf{19.4} \\ 28.6} &\makecell{19.7 \\ 28.6}  \\
   \bottomrule[1pt]
\end{tabular}
\vspace{1em}
\label{app:hyper}
\end{table}

\begin{table}[ht]
\vspace{-1em}
\caption{Ablation results of IPMix on CIFAR-100 with WRN40-4. While the addition + multiplication achieves the highest accuracy, it compromises corruption and calibration. In contrast, random mixing operations bolster robustness and calibration. Experiment results demonstrate that combining all mixing operations achieves the most balanced performance.}
\vspace{0.5em}
\centering
\begin{tabular}{cccc}
    \toprule
     \multirow{2}{*}{Mixing operations}& \multicolumn{1}{c}{Classification} & \multicolumn{1}{c}{Robustness} & \multicolumn{1}{c}{Calibration} \\
    &Error($\downarrow$) &mCE($\downarrow$) &RMS($\downarrow$)\\
    \midrule
    Addition + Multiplication   &\textbf{19.2} & 31 & 4.1   \\ 
    Random pixels mixing   &19.6 & 28.7 & 3.7  \\
    Random elements mixing   &19.9 & 28.8 & \textbf{2.7}  \\
    IPMix   &19.4 & \textbf{28.6} & 2.8  \\
   \bottomrule[1pt]
\end{tabular}
\label{app:operations}
\end{table}

\begin{table}[ht!]
\caption{The results of patch variants ablation on CIFAR-100 with ResNeXt-29. }
\vspace{0.5em}
\centering
\begin{tabular}{cccc}
    \toprule
    \multirow{2}{*}{Variants}& \multicolumn{1}{c}{Classification} & \multicolumn{1}{c}{Robustness} & \multicolumn{1}{c}{Calibration} \\
    &Error($\downarrow$) &mCE($\downarrow$) &RMS($\downarrow$)\\
    \midrule
    IPMix-Square   &\textbf{18.3} & 28.5 & 3.9   \\ 
    IPMix-Scar   &18.6 & \textbf{28.0} & 4.1  \\
    IPMix   &\textbf{18.3} & 28.1 & \textbf{3.8}  \\
   \bottomrule[1pt]
\end{tabular}
\label{app:variants}
\end{table}

\subsection{Additional Robustness Experiments}
\label{app:add_robust}
Recent works propose that some data augmentation techniques are tailored to particular datasets when testing model robustness. To evaluate the generality of IPMix, we experiment with other types of distribution shifts beyond common corruptions. We examine IPMix on CIFAR-10-{$\overline{\mbox{C}}$}, CIAFR-100-{$\overline{\mbox{C}}$}, and ImageNet-{$\overline{\mbox{C}}$}~\cite{Mintun2021OnIB}. CIFAR-10-{$\overline{\mbox{C}}$}, CIFAR-100-{$\overline{\mbox{C}}$}, and ImageNet-{$\overline{\mbox{C}}$} are similar to CIFAR-C and ImageNet-C but utilize a different set of corruptions.Results in Table~\ref{extent} demonstrate that IPMix achieves SOTA or comparison results by comparing with other methods.
 
\begin{table}[ht]
\caption{Results of robustness resist other distribution shifts. Bold is best.}
\vspace{1em}
\centering
\begin{tabular}{ccccc}
    \toprule
    Methods& CIFAR-10-{$\overline{\mbox{C}}$} & CIFAR-100-{$\overline{\mbox{C}}$} & ImageNet-{$\overline{\mbox{C}}$}\\
    \midrule
    Vanilla    &26.4 & 52  &60.2 \\ 
    MixUp~\cite{Zhang2017mixupBE}    &22.4 & 50  &54.1 \\ 
    CutOut~\cite{Devries2017ImprovedRO}    &24.2 &50.1 & 58.4   \\ 
    CutMix~\cite{Yun2019CutMixRS}    &25.1 &49.9 & 57.8   \\ 
    AugMix~\cite{Hendrycks2019AugMixAS}   &19.3 &41 & 54.3  \\
    PixMix~\cite{Hendrycks2021PixMixDP}  &13.6 &36.7 & \textbf{47.1}  \\
    IPMix  &\textbf{13} & \textbf{36} & 47.9\\
   \bottomrule[1pt]
\end{tabular}
\label{extent}
\end{table}

To better assess the performance of IPMix against natural distribution shifts, we extended our evaluation to various ImageNet benchmarks. We test IPMix on ObjectNet~\cite{Barbu2019ObjectNetAL}, ImageNet-E~\cite{Li2023ImageNetEBN}, ImageNet-Sketch~\cite{Wang2019LearningRG}, ImageNet-V2~\cite{Geirhos2018ImageNettrainedCA}, and Stylized-ImageNet~\cite{Recht2019DoIC}. The results presented in Table~\ref{shifts} indicate that IPMix consistently outperforms under diverse data shifts, underscoring its capability to enhance model robustness.

\begin{table}[ht]
\caption{Results of IPMix against natural distribution shifts. Higher is better.}
\vspace{0.5em}
\centering\resizebox{\textwidth}{!}{
\begin{tabular}{cccccc}
    \toprule
    & ObjectNet & ImageNet-E & ImageNet-Sketch &ImageNet-V2 &Stylized-ImageNet\\
    \midrule
    Vanilla    &17.3 & 76.7  &24.2 & 63.3  &7.4\\ 
    MixUp~\cite{Zhang2017mixupBE}    &18.4 & 77.1  &24.4 & 63.6  &7.3\\ 
    CutOut~\cite{Devries2017ImprovedRO}    &17.3 &24.1 & 58.4  & 63.7  &7.6 \\ 
    CutMix~\cite{Yun2019CutMixRS}    &18.9 &76.7 & 23.8 & 65.4  &5.3  \\ 
    AugMix~\cite{Hendrycks2019AugMixAS}   &17.6 &78.6 & 28.5 & 65.2  &11.2 \\
    PixMix~\cite{Hendrycks2021PixMixDP}  &18.5 &80 & 29.2 & \textbf{65.8}  &11.8 \\
    IPMix  &\textbf{19.3} & \textbf{80.9} & \textbf{31.1} & 65.6  &\textbf{12.2}\\
   \bottomrule[1pt]
\end{tabular}}
\label{shifts}
\end{table}

\section{Evaluation Metrics}
\label{app:eva}
We evaluate various safety measures on CIFAR and ImageNet, including corruption robustness, calibration, adversarial robustness, consistency, and anomaly detection. Task evaluation metrics are shown below.

\textbf{Corruption robustness.} Following AugMix, we utilize the Mean Corruption Error (mCE) to test a model's resistance to corrupted data on CIFAR-10-C, CIFAR-100-C, and ImageNet-C. Mean Corruption Error is the mean error rate normalized by the corruption errors of a baseline model over 15 corruption types and 5 corruption severity.  We train AlexNet~\cite{Krizhevsky2012ImageNetCW} as the baseline for ImageNet experiments. 

\textbf{Calibration.}
The calibration task is to verify whether the predicted probability estimates are representative of the true correctness likelihood. We use RMS Calibration Error~\cite{Hendrycks2018DeepAD} as the metric, which can be computed as $\sqrt{\mathbb{E}_C[(\mathbb{P}(Y = \hat{Y}|C=c)-c)^{2}]}$, where $C$
is the classifier’s confidence that its prediction $\hat{Y}$ is correct. Lower is better.

\textbf{Adversarial robustness.}
We utilize PGD to verify the adversarial robustness of image classifiers. We use 20 steps of optimization and an $\ell_\infty$ budget of 2/255 on CIFAR-10 and CIFAR-100. The metric is the classifier error rate. Lower is better.

\textbf{Consistency.}
Following AugMix, we verify perturbation consistency on CIFAR-10-P, CIFAR-100-P, and ImageNet-P. The metric is the mean flip rate (mFR), which can be tested through video frame predictions normalized by a baseline model matched by 10 different perturbation types. We choose AlexNet as the baseline model.

\textbf{Anomaly detection.}
We utilize two challenging datasets, ImageNet-A and ImageNet-O to evaluate model robustness under out-of-distribution shifts. The main metric on ImageNet-A is accuracy,  and on ImageNet-O is the area under the precision-recall curve (AUPR). Higher is better. The anomaly score is the negative of the maximum softmax probabilities~\cite{Hendrycks2016ABF}.

\section{The Algorithm of IPMix}
\label{app:algo_IPMix}
The algorithm to generate IPMix images is summarized in Algorithm 
 \ref{app:algo}. The fractals we use are selected at random from the IPMix fractal set (for further details, please see Appendix~\ref{app:fractal}). On CIFAR, the patch sizes we employ are randomly chosen from a set including 4, 8, 16, and 32, whereas for ImageNet-1K, we opt for patch sizes from 4, 8, 16, 32, 64, and 256. We randomly mix the augmented original image to increase diversity. Across all our experiments, we consistently use $k=3$ and $t=3$.

\begin{algorithm}
\DontPrintSemicolon
\SetKwData{Left}{left}\SetKwData{This}{this}\SetKwData{Up}{up}
\SetKwFunction{Union}{Union}\SetKwFunction{FindCompress}{FindCompress}
\SetKwInOut{Input}{input}\SetKwInOut{Output}{output}

\Input{Origin image $x$, fractal $x_{\text{fractal}}$, augmentation methods $M$=\{image-level, P-level\}, patch sizes $P_{\text{size}}$ , P operations $P$ = \{random pixels mixing,...,add\}, image operations $I$ = \{invert,...,mirror\} , width $k$, max depth $t$.}
\Output{$x_{\text{IPMix}}$}
Sample mixing weights $w_1$,...,$w_k$ $\sim $ Dirichlet($\alpha$,...,$\alpha$)\\
Sample weights $m$ $\sim $ Beta($\alpha$,$\alpha$) \\
Generate $x_{\text{mix}}$ = Zerolikes($x$) \\

\For{$i \leftarrow 1$ \KwTo $k$}{
    Generate $x_{\text{mixed}} = x.\text{copy}()$\\
    Randomly choose method $\text{'meth'}$ from $M$\\
    \If{$\normalfont \text{'meth'}$ == $\normalfont \text{'P-level'}$}{
        \For{$j=1$ {\bfseries to} $\normalfont \text{random.choose([1,...,$t$])}$}{
           Random sample size $s$ from $P_{\text{size}}$ \tcp*{$P_{\text{size}} = \text{x.size} \rightarrow \text{Pixel-level op}$}

           Sample operations $p_o$ from $P$ \\
           \If{$\normalfont \text{random.random()}$ $>$ 0.5}{
           $x_{\text{mixed}}$ = patch mixing($x_{\text{mixed}}$, $x_{\text{fractal}}$, $s$, $p_o$) \tcp*{ See Sec.4.2}
           }                           
           \Else{
             Sample operations $i_o$ from $I$  \tcp*{ For diversity increase}
             $x_{\text{aug}}$ = $i_o$($x$)\\
             $x_{\text{mixed}}$ = patch mixing($x_{\text{mixed}}$, $x_{\text{aug}}$, $s$, $p_o$)
           }
        }
    }
    \Else{
        \For{$j=1$ {\bfseries to} $\normalfont \text{random.choose([1,...,$t$])}$}{
           Sample operations $i_o$ from $I$ \\
           $x_{\text{mixed}}$ = $i_o$($x_{\text{mixed}}$)
        }
    }
    $x_{\text{mix}}$ += $w_i$ $\cdot$ $x_{\text{mixed}}$ \tcp*{ $w_i$ from Dirichlet($\alpha$,...,$\alpha$)}
}
\KwRet $x_{\text{IPMix}} = m \cdot x_{\text{mix}} + (1 - m) \cdot x$ \tcp*{ $m$ from Beta($\alpha$,$\alpha$)}

\caption{Generate IPMix Images}\label{app:algo}
\end{algorithm}

\section{Generating Fractal Images}
\label{app:fractal}
While prior works have exclusively utilized Iterated Function Systems (IFS) to generate fractal data \cite{Kataoka2021PreTrainingWN, Anderson2021ImprovingFP}, various other fractal-generating programs can also be employed. To further enhance the structural complexity and diversity, we have ventured beyond IFS and incorporated the Escape-time Algorithm to generate 'orbit trap' complex fractals. The most common 'orbit trap' fractal images, Mandelbrot and Julia fractals, can be derived from Eq. (\ref{fractals}):
\begin{equation} F(z) = z^{2} + c \label{fractals} \end{equation}
In Eq. (\ref{fractals}), $z$ represents a complex number, and $c$ is a constant value. In the case of the Mandelbrot set, we initialize $z$ at 0, with $c$ corresponding to the specific coordinate in the complex plane that is under examination. Conversely, when generating the Julia set, $c$ remains constant throughout the set, and $z$ is initiated as the particular coordinate that is currently being tested.

Moreover, guided by the approach of \cite{Anderson2021ImprovingFP}, we create an additional 3000 fractals, each rendered with a unique, randomly generated background and color scheme using IFS. Furthermore, we supplement our dataset with an additional fractals obtained from DeviantArt\footnote{\url{https://www.deviantart.com/}}. These images, exhibiting greater complexity than those generated via IFS or the Escape-time Algorithm, significantly enhance dataset diversity. Besides, we collect 4000 feature images to improve diversity. In total, we assemble a collection of 13000 images named IPMix set for increasing data diversity and structural complexity when mixed with clean images.

\section{The Details about Combination Experiments}
\label{app:combination}
In section~\ref{section:3}, we show that simply combining different levels of approaches can degrade model performance across various metrics. Building upon these findings, in this part, we want to examine the impact of the order of operations on combination experiments.

In our experiments, we adopt MixUp~\cite{Zhang2017mixupBE}, CutMix~\cite{Yun2019CutMixRS}, and AugMix~\cite{Hendrycks2019AugMixAS} as representative techniques for pixel-level, patch-level, and image-level augmentation, respectively. In all experiments, we apply AugMix first, followed by CutMix or MixUp. The rationale behind this order is that AugMix is commonly used in PIL images to enhance data diversity. In contrast, MixUp and CutMix interpolate and mix images after images conversion into tensors. Furthermore, applying Mixup/CutMix before AugMix could lead to unnatural transformations, as AugMix operations would distort the mixed images, counteracting the aim of preserving the individual image context during interpolation.

We have adopted several different combinations as follows.
\begin{itemize}
\item First, we apply AugMix, then MixUp, and finally CutMix.
\item First, we apply AugMix, then CutMix, and finally MixUp.
\item We apply AugMix first, followed by either CutMix or MixUp, chosen randomly.
\item We apply AugMix first. Depending on the training epochs, we use either CutMix or MixUp.
\end{itemize}

\begin{table}[ht]
\vspace{-1em}
\caption{The combination experiments of different levels of data augmentation on CIFAR-100.}
\vspace{1em}
 \centering
\begin{tabular}{cccc}
    \toprule
    \multirow{2}{*}{Methods} & \multicolumn{1}{c}{Classification} & \multicolumn{1}{c}{Robustness} & \multicolumn{1}{c}{Calibration} \\
    &Error($\downarrow$) &mCE($\downarrow$) &RMS($\downarrow$)\\
    \midrule
    Vanilla  &21.3 &50  &14.6  \\ 
    MixUp    &20.5 &45.9  &10.5  \\ 
    CutMix    &20.3 &50  &9.3  \\ 
    AugMix    &20.6 &33.3  &6.3  \\ 
    AugMix$\rightarrow$MixUp$\rightarrow$CutMix & 23.4 & 50.1& 25.6\\
    AugMix$\rightarrow$CutMix$\rightarrow$MixUp & 27 & 51.4& 26.7\\
    Chosen Randomly ($p=0.5$)  & 22.6  & 40.6 & 19 \\
    Epoch-Dependent  & 21.1 & 37.6 & 7.2 \\
   \bottomrule[1pt]
\end{tabular}
\label{app:comb}
\end{table}

In all experiments, we use the optimal hyperparameters specified in the original papers. We set $k=3$ for AugMix and $\alpha=1$ for MixUp and CutMix. The results are demonstrated in Table~\ref{app:comb}.

We set the total number of training epochs to 100 on 40-4 Wide ResNet for all experiments. In our Epoch-Dependent combination experiments, we found that employing MixUp for the initial 50 epochs and transitioning to CutMix for the rest yielded the best performance. Nevertheless, it doesn't perform as well as the individual augmentation techniques. This underperformance might be due to the increased complexity in the synthesized training instances, possibly impeding the extraction of discriminative feature representations by models. Further experiments could explore different combinations of these techniques to improve their effectiveness. 

In order to thoroughly analyze the influence of the augmentation strength of each method, we have conducted experiments considering various hyperparameter combinations. Specifically, we evaluated $k$ = 1, 3, 5 (for AugMix) and $\alpha $ = 0.2, 0.5, 1 (for MixUp and CutMix). We opted to exclude $k$ = 3 and $\alpha$ = 1, the original optimal hyperparameters in their papers, thereby reducing the total combinations from 27 to 8. From the experimental results in Table~\ref{app:operas}, combining different hyperparameters does not significantly improve the model performance. We set the total number of training epochs to 100 for all experiments with WRN40-4 on CIFAR-100.

\begin{table}[ht]
\vspace{-1em}
\caption{Could decreasing the augmentation strength of each method yield better performance?}
\vspace{1em}
\centering
\begin{tabular}{cccc}
    \toprule
     \multirow{2}{*}{Combination }& \multicolumn{1}{c}{Classification} & \multicolumn{1}{c}{Robustness} & \multicolumn{1}{c}{Calibration} \\
    &Error($\downarrow$) &mCE($\downarrow$) &RMS($\downarrow$)\\
    \midrule
    $\alpha=0.2$,$\alpha=0.2$,$k=1$   &23.9 & 51.2 & 25.3  \\
    $\alpha=0.2$,$\alpha=0.2$,$k=5$   &24.5 & 51 & 25.3  \\
    $\alpha=0.2$,$\alpha=0.5$,$k=1$   &26 & 50.7 & 24.9  \\
    $\alpha=0.2$,$\alpha=0.5$,$k=5$   &24.4 & 50.6 & 25.7  \\
    $\alpha=0.5$,$\alpha=0.2$,$k=1$   &25.8 & 50.8 & 25.4  \\
    $\alpha=0.5$,$\alpha=0.2$,$k=5$   &25 & 49.1 & 24.8  \\
    $\alpha=0.5$,$\alpha=0.5$,$k=1$   &25.5 & 50.5 & 25.1  \\
    $\alpha=0.5$,$\alpha=0.5$,$k=5$   &26 & 51.2 & 25.9  \\
   \bottomrule[1pt]
\end{tabular}
\label{app:operas}
\end{table}

\section{Training Time}
\label{app:time}
In this section, we present a comparative analysis of the training time. The results in Table~\ref{traing_time} show that IPMix adds only a modest training overhead over Vanilla, which is advantageous for its practical use in real-world scenarios. 

\begin{table}[ht]
\caption{We test IPMix on two NVIDIA GeForce RTX 3090 GPUs with ResNet18 for 90 epochs. The training time of IPMix is acceptable by comparison with other data augmentation methods. }
\vspace{1em}
\centering
\begin{tabular}{cc}
    \toprule
    Method & Time(sec/epochs) \\
    \midrule
    Vanilla  &3764    \\ 
    MixUp~\cite{Zhang2017mixupBE}     &3913   \\ 
    CutOut~\cite{Devries2017ImprovedRO}    &3870   \\ 
    CutMix~\cite{Yun2019CutMixRS}    &4139   \\ 
    AugMix~\cite{Hendrycks2019AugMixAS}    &4762   \\ 
    PixMix~\cite{Hendrycks2021PixMixDP}    &4310   \\ 
    AugMax~\cite{Wang2021AugMaxAC}    &7564    \\
    IPMix  &4380    \\ 
   \bottomrule[1pt]
\end{tabular}
\label{traing_time}
\end{table}

\section{Full Results of IPMix across Architectures}
\label{app:full_res}
In Table \ref{app:full1}, we show the full results of IPMix across architectures on CIFAR-10 and CIFAR-100. 
\begin{table*}[ht]
\caption{Full results for IPMix on CIFAR-10 and CIFAR-100. We test the ability of IPMix on accuracy, robustness, consistency, adversaries, and calibration across different models. Top: CIFAR-10. Bottom : CIFAR-100.}
\centering
\begin{tabular}{cccccccc}
    \toprule
    & \multicolumn{1}{c}{Classification} & \multicolumn{1}{c}{Robustness} & \multicolumn{1}{c}{Consistency} & \multicolumn{1}{c}{Adversaries} & \multicolumn{1}{c}{Calibration} \\
    &Error($\downarrow$) &mCE($\downarrow$) &mFR($\downarrow$) &Error($\downarrow$) &RMS($\downarrow$) \\
    \midrule
    WideResNet40-4    &4 & 8.6 & 1.3 & 74.4 & 2.3 \\ 
    WideResNet28-10   &3.3 & 7.5 & 1.1 & 76.4 & 1.9 \\
    ResNeXt-29         &3.8 & 8.6 & 1.4 & 93.2 & 2 \\
    ResNet-18          &4.2 & 8.4 & 1.7 & 80 & 2.4 \\
    \midrule
    Mean               &3.8 & 8.3 & 1.4 & 81 & 2.2 \\
   \midrule
    WideResNet40-4    &19.4 & 28.6 & 4.3 & 89.4 & 2.8 \\ 
    WideResNet28-10  &17.4 & 26.6 & 4.2 & 91.3 & 6.4 \\
    ResNeXt-29         &18.3 & 28.1 & 5 & 96.9 & 3.8\\
    ResNet-18          &21.6 & 29.9 & 5.4 & 95.6 & 6.3\\
    \midrule
    Mean               &19.2 & 28.3 & 4.7 & 93.3 & 4.9\\
   \bottomrule
\end{tabular}
\label{app:full1}
\end{table*}

\section{More CAM Visualizations}
\label{app:CAM}
\begin{figure}[t]
\vspace{-0.5em}
\begin{center}
\centerline{\includegraphics[width=0.95\columnwidth]{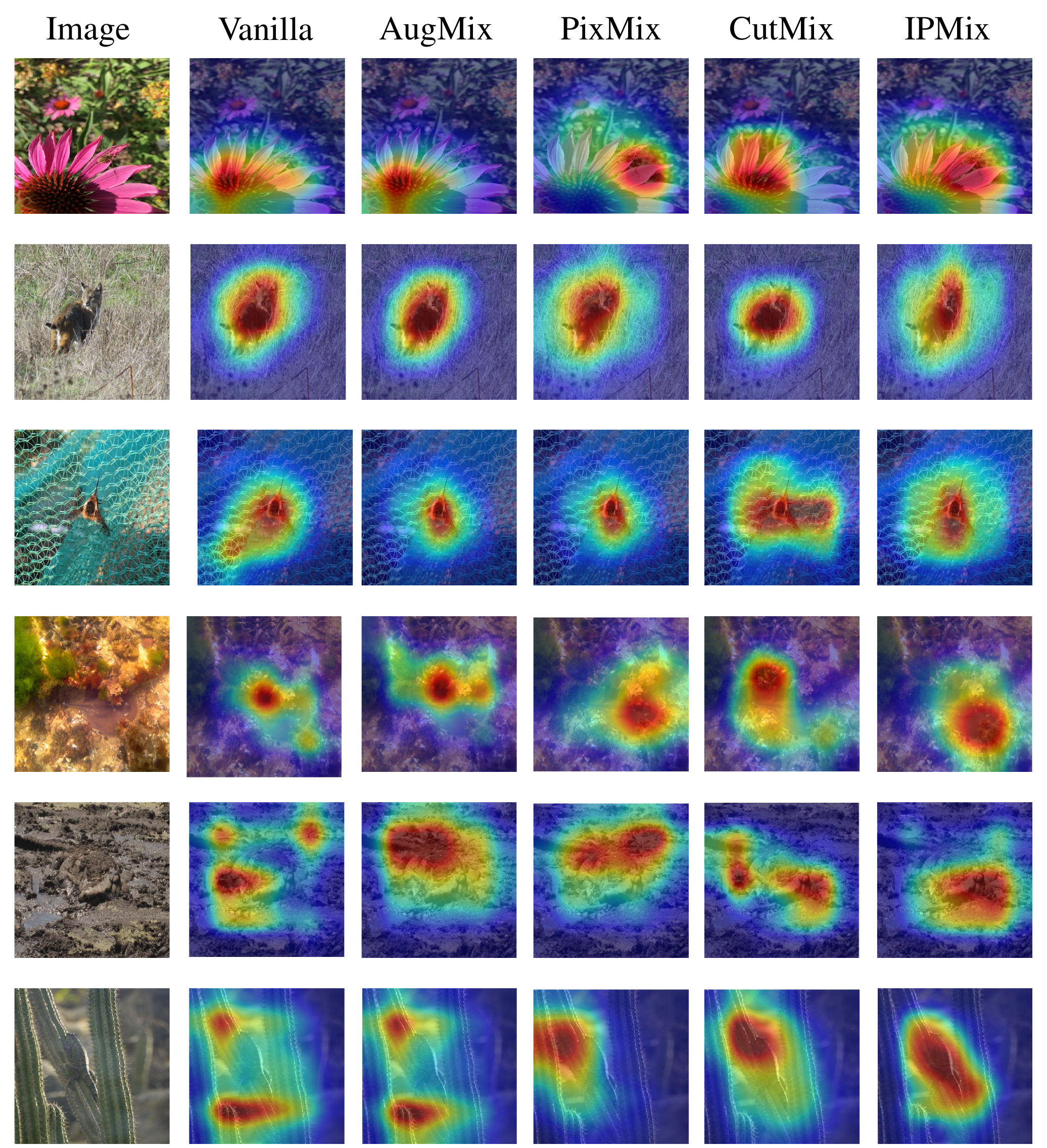}}
\vspace{1em}
\caption{More CAM visualizations of IPMix. Input images come from ImageNet-A, the most challenging dataset to verify the performance
of model classifiers against distribution shifts.}
\label{app:CAM_full}
\end{center}
\end{figure}

In this section, we demonstrate more CAM visualizations of IPMix, as shown in Figure~\ref{app:CAM_full}.

\section{The Analysis of Ablation Experiments}
\label{app:analysis}

\begin{table}[ht]
        \caption{Ablation results of different components of IPMix on CIFAR-100. Mean and standard derivation over three random seeds is shown for each experiment. Bold is the best.}
        \vspace{0.5em}
    \centering
    \begin{tabular}{cccccc}
        \toprule
    & \multicolumn{1}{c}{Classification} & \multicolumn{1}{c}{Robustness} & \multicolumn{1}{c}{Consistency} & \multicolumn{1}{c}{Adversaries} & \multicolumn{1}{c}{Calibration} \\
    &Error($\downarrow$) &mCE($\downarrow$) &mFR($\downarrow$) &Error($\downarrow$) &RMS($\downarrow$) \\
    \midrule
        IPMix &\textbf{19.4$_{(\pm0.17)}$} &\textbf{28.6$_{(\pm0.2)}$} & 89.4$_{(\pm0.18)}$ &\textbf{4.3$_{(\pm0.09)}$} &\textbf{2.8$_{(\pm0.07)}$}  \\ 
        \midrule
        w/o patch  &19.7$_{(\pm0.13)}$ & 30 $_{(\pm0.21)}$ & 91.7 $_{(\pm0.15)}$ &4.7 $_{(\pm0.02)}$ &4.6 $_{(\pm0.07)}$\\
        w/o pixel  &19.6 $_{(\pm0.09)}$ & 33 $_{(\pm0.35)}$ & 92.6 $_{(\pm0.20)}$ & 5.2 $_{(\pm0.05)}$ & 8.2 $_{(\pm0.12)}$ \\
        w/o image  &20.1 $_{(\pm0.27)}$ & 34 $_{(\pm0.65)}$ & \textbf{87.8 $_{(\pm0.22)}$} &5.5 $_{(\pm0.11)}$ &8.6 $_{(\pm0.21)}$\\
       \bottomrule[1pt]
    \end{tabular}
    \vspace{0em}
    \label{app:ablation_anlysis}
\end{table}

In this section,  we will detailed analyze the impact of each part on different safety metrics through ablation experiment results shown in Table~\ref{app:ablation_anlysis}.

\textbf{Accuracy}: The image-level augmentation has the most substantial effect on accuracy, aligning with current findings~\cite{Cubuk2019AutoAugmentLA,Cubuk2019RandaugmentPA} that image-level methods are commonly used to boost accuracy.

\textbf{Robustness}: Both pixel-level and image-level augmentations improve robustness. Since pixel-level introduces fine-grained variations for pattern recognition, while image-level increases dataset diversity, preventing the model from merely memorizing fixed augmentations.

\textbf{Calibration and Consistency}: The Image-level part significantly influences calibration and consistency, which increases diversity to improve the prediction calibration across scenarios and ensures consistency in responses to minor perturbations.

\textbf{Adversarial Attacks}: Without the image-level component, adversarial performance improves, implying diverse data might \textbf{weaken} defense against attacks. Conversely, removing pixel-level methods will degrade adversarial robustness, given their inherent resistance to perturbations.

\section{The Experiment Results on Transformer Architecture}
\label{app:Vit}
In this section, we will evaluate the performance of IPMix on Vision Transformer. We trained a small ViT for 300 epochs on CIFAR-10 and CIFAR-100. This step aimed to confirm IPMix's potential on smaller datasets using Transformer architectures. In future work, we plan to expand our experiments with transformer architectures.
The experiment results in Table~\ref{app:vit_exem} and Table~\ref{app:vit_exem_2} show that IPMix achieves the best performance on ViT.

\begin{table}[ht]
        \caption{Experiments on CIFAR-10. Bold is the best.}
        \vspace{0.5em}
    \centering
    \begin{tabular}{cccccc}
        \toprule
    & \multicolumn{1}{c}{Classification} & \multicolumn{1}{c}{Robustness} & \multicolumn{1}{c}{Consistency} & \multicolumn{1}{c}{Adversaries} & \multicolumn{1}{c}{Calibration} \\
    &Error($\downarrow$) &mCE($\downarrow$) &mFR($\downarrow$) &Error($\downarrow$) &RMS($\downarrow$) \\
        \midrule
        Vanilla & 19.5$_{(\pm 0.07)}$ & 27.7$_{(\pm 0.14)}$ & 91.3$_{(\pm 0.13)}$ & 5.9$_{(\pm 0.02)}$ & $10$ \\
        MixUp & 1$_{(\pm 0.11)}$ & 34.7$_{(\pm 0.21)}$ & 89.3$_{(\pm 0.21)}$ & 6$_{(\pm 0.05)}$ & 9.9$_{(\pm 0.03)}$ \\
        CutMix & 19.3$_{(\pm 0.08)}$ & 34.3 $_{(\pm 0.19}$ & 89.1$_{(\pm 0.14)}$ & 5.5$_{(\pm 0.05)}$ & 7.5 $_{(\pm 0.02)}$ \\
        PixMix & 28.4$_{(\pm 0.14)}$ & 33.$_{(\pm 0.24)}$ & 91$_{(\pm 0.12)}$ & 6.5$_{(\pm 0.11)}$ & 4.4$_{(\pm 0.07)}$ \\
        AugMix & 20.3$_{(\pm 0.14)}$ & 25.6$_{(\pm 0.2)}$ & 80.3$_{(\pm 0.16)}$ & 5.1$_{(\pm 0.09)}$ & 6$_{(\pm 0.08)}$\\
        IPMix & \textbf{19.2$_{(\pm 0.12)}$} & \textbf{23.7$_{(\pm 0.2)}$} & \textbf{75.8$_{(\pm 0.13)}$} & \textbf{3.7$_{(\pm 0.07)}$} & \textbf{5.3$_{(\pm 0.07)}$} \\
       \bottomrule[1pt]
    \end{tabular}
    \vspace{0em}
    \label{app:vit_exem}
\end{table}

\begin{table}[ht]
        \caption{Experiments on CIFAR-100. Bold is the best.}
        \vspace{0.5em}
    \centering
    \begin{tabular}{cccccc}
        \toprule
    & \multicolumn{1}{c}{Classification} & \multicolumn{1}{c}{Robustness} & \multicolumn{1}{c}{Consistency} & \multicolumn{1}{c}{Adversaries} & \multicolumn{1}{c}{Calibration} \\
    &Error($\downarrow$) &mCE($\downarrow$) &mFR($\downarrow$) &Error($\downarrow$) &RMS($\downarrow$) \\
        \midrule
        Vanilla & 40.1$_{(\pm 0.12)}$ &  56.3$_{(\pm 0.1)}$ & 96.2$_{(\pm 0.14)}$ & 12.4$_{(\pm 0.04)}$ &  14.8$_{(\pm 0.02)}$ \\
        MixUp & 40$_{(\pm 0.14)}$ &  56 $_{(\pm 0.18}$ &  92.5$_{(\pm 0.18)}$ &  9.8$_{(\pm 0.03)}$ & 9.5 $_{(\pm 0.02)}$ \\
        CutMix & 39.5$_{(\pm 0.11)}$ &  56.3 $_{(\pm 0.15}$ &  96.2$_{(\pm 0.17)}$ &  10$_{(\pm 0.03)}$ &  9.8 $_{(\pm 0.03)}$ \\
        PixMix &48.7$_{(\pm 0.14)}$ & 54.3$_{(\pm 0.21)}$ & 93.2$_{(\pm 0.14)}$ & 10.9$_{(\pm 0.17)}$ &  4.9$_{(\pm 0.04)}$ \\
        AugMix & 35.3$_{(\pm 0.17)}$ & 42.4$_{(\pm 0.21)}$ &  84.6$_{(\pm 0.16)}$ & 6.9$_{(\pm 0.03)}$ &  6.4$_{(\pm 0.07)}$\\
        IPMix & \textbf{32.6$_{(\pm 0.11)}$} &  \textbf{39.6$_{(\pm 0.23)}$} & \textbf{83.2$_{(\pm 0.15)}$} & \textbf{6.3$_{(\pm 0.04)}$} & \textbf{5.3$_{(\pm 0.05)}$} \\
       \bottomrule[1pt]
    \end{tabular}
    \vspace{0em}
    \label{app:vit_exem_2}
\end{table}

\section{The Drawbacks of Different Levels of Methods}
\label{app:drawbacks}
In this section, we will reveal the drawbacks of different levels of approaches and explain how IPMix solves these problems.

The drawbacks of label variant methods:

\textbf{Pixel-level:} Mixing images with distinct labels and linearly interpolating between them will impose certain “local linearity” constraints on the model’s input space beyond the data manifold, which may lead to "manifold intrusion". Consider one experiment on MNIST. If we use MixUp to linearly mix two numbers, such as "1" and "5", the generated image will show the characteristics of "8". When the generated "8" collides with a real "8" in the data manifold, there will be a problem of manifold intrusion. Since the two samples have similar characteristics, one is the real label and the other is a soft label ("1" and "5"). This will interfere with its ability to understand and classify categories and degrade model performance.

\textbf{Patch-level:} The problem of manifold intrusion also occurs in the patch-level method, termed "label mismatch." This occurs when the chosen source patch doesn't accurately represent the source object, leading the interpolated label misleads the model to learn unexpected feature representation. For example, using CutMix to mix images of a cat and a dog. CutMix might select 20 \% of the background area from the cat image without information about the object (cat). However, their interpolated labels encourage the model to learn both objects’ features (dog and cat) from that training image and degrade model performance.

The drawbacks of image-level methods:

\textbf{Image-level} data augmentation increases data diversity by applying label-preserving transformations to the whole image. Notable among these are search-based methods like AutoAugment, RandAugment, and FastAugment. While they improve performance effectively, the computationally expensive search for an optimal augmentation policy often exceeds the training process’s complexity. Thus, efforts to minimize the search space, optimize search parameters, and uncover potential universal pipelines are central to the effectiveness of these methods.

In conclusion, we solve these questions by:

\begin{itemize}[leftmargin=0.5cm, itemindent=0cm]
    \item  Incorporate structural complexity from synthetic data at various levels to produce more diverse images. Our method is \textbf{label-preserving}, ensuring it is not affected by manifold intrusion.
    \item  Randomly sample operations from PIL (e.g., brightness, sharpness) and randomly sample strengths to enhance the diversity of training data \textbf{without expensive searching}.
    \item Integrate three levels of data augmentation into a single framework with limited computational overhead, demonstrating that these approaches are complementary and that a unification among them is necessary to achieve robustness.
\end{itemize}

\section{Limitation and Broader Impact}
\label{app:limit}

While IPMix has shown promising results, the theoretical foundation of IPMix requires further development to gain deeper insights into its underlying principles. Meanwhile, our approach primarily focuses on CNN, and its effectiveness on Visual Transformers requires additional experimental validation. Additionally, the experiments conducted on a limited set of safety metrics, and the performance of IPMix in real-world scenarios with more comprehensive safety measures warrants future investigation~\cite{Hendrycks2021UnsolvedPI}. In continuing our efforts to refine and enhance the IPMix methodology, we will focus on addressing these limitations in future works.

Since IPMix improves various safety measures, it can generate many beneficial effects in real-world environments, improving the robustness against attacks and the calibrated prediction confidence of models. Moreover, IPMix integrates three levels of data augmentation into a single framework, demonstrating that these approaches are complementary and necessary to achieve better performance. We believe the improvements in safety metrics and the coherent framework of combining various techniques will shed light on this field.
%%%%%%%%%%%%%%%%%%%%%%%%%%%%%%%%%%%%%%%%%%%%%%%%%%%%%%%%%%%%
\end{document}